\documentclass[lettersize,journal]{IEEEtran}

\usepackage{amsmath,amsfonts}
\usepackage{bm}
\usepackage{makecell}

\usepackage[ruled,vlined,linesnumbered]{algorithm2e}
\usepackage{algpseudocode}

\usepackage{array}
\usepackage{booktabs}
\usepackage{multirow}

\usepackage[caption = false,font = normalsize,labelfont = sf,textfont = sf]{subfig} 
\usepackage{graphicx}
\usepackage{stfloats} 

\usepackage[table,xcdraw]{xcolor}
\usepackage{utfsym} 
\definecolor{mydeepblue}{RGB}{0,0,255}
\definecolor{mydeepgreen}{RGB}{50,205,50}
\newcommand{\bluecross}{\textcolor{mydeepblue}{\usym{2718}}} 
\newcommand{\greencheck}{\textcolor{mydeepgreen}{\usym{2714}}} 

\usepackage{textcomp}
\usepackage{url}
\usepackage{microtype}
\usepackage{tikz}
\usepackage{verbatim}
\usepackage{orcidlink}
\usepackage[nocompress]{cite} 

\usepackage[normalem]{ulem} 

\usepackage{hyperref} 
\hypersetup{
 colorlinks = true,
 linkcolor = blue,
 filecolor = magenta,
 citecolor = blue,
 urlcolor = black
}
\usepackage[capitalise]{cleveref} 

\begin{document}

\title{PAD: Phase-Amplitude Decoupling Fusion for Multi-Modal Land Cover Classification}

\author{%
Huiling~Zheng\orcidlink{0009-0002-5095-7575},
Xian~Zhong\orcidlink{0000-0002-5242-0467},~\IEEEmembership{Senior~Member,~IEEE},
Bin~Liu\orcidlink{0000-0001-9873-6160},~\IEEEmembership{Member,~IEEE},
Yi~Xiao\orcidlink{0000-0001-9533-8917},~\IEEEmembership{Member,~IEEE},
Bihan~Wen\orcidlink{0000-0002-6874-6453},~\IEEEmembership{Senior~Member,~IEEE},
and~Xiaofeng~Li\orcidlink{0000-0001-7038-5119},~\IEEEmembership{Fellow,~IEEE}%

\thanks{Manuscript received April 21, 2025; revised July 3 and September 27, 2025; accepted October 7, 2025. This work was supported in part by the National Natural Science Foundation of China (Grants No. 62271361 and 42006159) and the Hubei Provincial Key Research and Development Program (Grant No. 2024BAB039). Numerical calculations were partly carried out on the Big Earth Data Cloud Service Platform. (\textit{Corresponding authors: Xian Zhong; Bin Liu.})}%

\thanks{Huiling Zheng is with the Sanya Science and Education Innovation Park, Wuhan University of Technology, Sanya 572025, China; the School of Computer Science and Artificial Intelligence, Wuhan University of Technology, Wuhan 430070, China; and the Key Laboratory of Ocean Circulation and Waves, Institute of Oceanology, Chinese Academy of Sciences, Qingdao 266071, China (e-mail: zhenghl@whut.edu.cn).}%

\thanks{Xian Zhong is with the Hubei Key Laboratory of Transportation Internet of Things, School of Computer Science and Artificial Intelligence, Wuhan University of Technology, Wuhan 430070, China, and with the State Key Laboratory of Maritime Technology and Safety, Wuhan University of Technology, Wuhan 430063, China (e-mail: zhongx@whut.edu.cn).}%

\thanks{Bin Liu is with the College of Oceanography and Ecological Science, Shanghai Ocean University, Shanghai 201306, China (e-mail: bliu@shou.edu.cn).}%

\thanks{Yi Xiao is with the School of Computer and Artificial Intelligence, Zhengzhou University, Zhengzhou 450001, China
(e-mail: yixiao@zzu.edu.cn).}

\thanks{Bihan Wen is with the Rapid-Rich Object Search Lab, School of Electrical and Electronic Engineering, Nanyang Technological University, Singapore 639798 (e-mail: bihan.wen@ntu.edu.sg).}%

\thanks{Xiaofeng Li is with the Key Laboratory of Ocean Circulation and Waves, Institute of Oceanology, Chinese Academy of Sciences, Qingdao 266071, China (e-mail: xiaofeng.li@ieee.org).}%

}

\markboth{IEEE Transactions on Geoscience and Remote Sensing, 2025}
{Shell \MakeLowercase{\textit{et al.}}: A Sample Article Using IEEEtran.cls for IEEE Journals}


\maketitle

\begin{abstract}

The fusion of Synthetic Aperture Radar (SAR) and RGB imagery for land cover classification remains challenging due to modality heterogeneity and underexploited spectral complementarity. Existing approaches often fail to decouple shared structural features from modality-complementary radiometric attributes, resulting in feature conflicts and information loss. To address this, we propose \textit{Phase-Amplitude Decoupling (PAD)}, a frequency-aware framework that separates phase (modality-shared) and amplitude (modality-complementary) components in the Fourier domain. This design reinforces shared structures while preserving complementary characteristics, thereby enhancing fusion quality. Unlike previous methods that overlook the distinct physical properties encoded in frequency spectra, PAD explicitly introduces amplitude-phase decoupling for multi-modal fusion. Specifically, PAD comprises two key components: \textit{1) Phase Spectrum Correction (PSC)}, which aligns cross-modal phase features via convolution-guided scaling to improve geometric consistency; and \textit{2) Amplitude Spectrum Fusion (ASF)}, which dynamically integrates high- and low-frequency patterns using frequency-adaptive multilayer perceptrons, effectively exploiting SAR’s morphological sensitivity and RGB’s spectral richness. Extensive experiments on \textsc{WHU-OPT-SAR} and \textsc{DDHR-SK} demonstrate state-of-the-art performance. This work establishes a new paradigm for physics-aware multi-modal fusion in remote sensing. The code will be available at \url{https://github.com/RanFeng2/PAD}.

\end{abstract}



\section{Introduction}

\IEEEPARstart{L}{and} cover classification (LCC) is a fundamental Earth observation task that categorizes surface materials, such as farmlands, forests, water bodies, and urban areas, using remote sensing imagery. This capability is essential for environmental monitoring, resource management, and urban development~\cite{liu2022env}. Recently, fusing Synthetic Aperture Radar (SAR) and RGB imagery has emerged as a promising approach to enhance LCC performance: SAR provides robust structural information resilient to atmospheric and illumination variations, while RGB imagery offers rich spectral details that mitigate SAR artifacts such as speckle noise~\cite{gao2024saropt, li2022review}. 

\begin{figure}[!t]
	\centering
	\includegraphics[width = \linewidth]{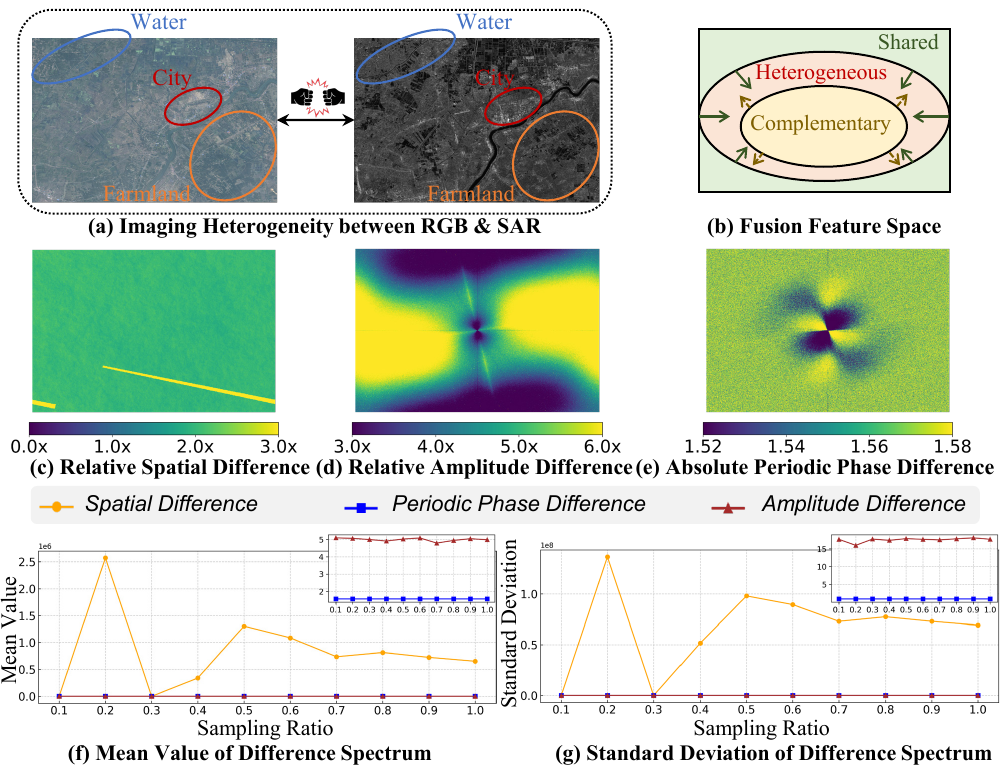}
	\caption{\textbf{Motivation and SAR-RGB Spatial-Frequency Difference Analysis.}
	(a) SAR and RGB imagery exhibit modality gaps due to distinct sensing mechanisms (backscattering \textit{vs.} reflectance). 
	(b) The fusion feature space contains heterogeneous, shared, and complementary components; effective fusion suppresses heterogeneous cues while enhancing shared and complementary patterns. 
	(c-g) Frequency-domain analysis on \textsc{WHU-OPT-SAR} (see \S\ref{sec:iii_a}): 
	(c) spatial discrepancies show anisotropic distributions; 
	(d) relative amplitude differences concentrate in high-frequency regions; 
	(e) phase discrepancies are minimal and primarily cluster in low-frequency bands; 
	(f-g) varying the sampling rate significantly affects spatial discrepancy distributions.}
	\label{fig1}
\end{figure}
 
Despite this complementarity, RGB-SAR fusion faces a key challenge: the inherent heterogeneity between microwave backscattering (SAR) and optical reflectance (RGB) causes spectral misalignment and hinders effective cross-modal feature integration (see \cref{fig1}(a)). In SAR images, objects often appear as discrete backscatter responses, lacking fundamental visual cues, such as geometric contours, textures, and consistent shapes, that are readily available in RGB imagery~\cite{zhou2025madinet}. Compared with natural image fusion, integrating SAR and RGB data presents pronounced radiometric and geometric disparities, which limits existing frameworks~\cite{ye2024optsarfusion, ma2024ftransUNet, xiao23mocg}. 

Deep learning, particularly convolutional neural networks (CNNs), has achieved remarkable success in LCC by extracting powerful features~\cite{wang2025balanced, liu2025mfae, xiao2022sr, zhong2021grayscale}. To mitigate modality heterogeneity, most approaches employ dual-branch architectures to extract multi-scale features independently and then fuse them via element-wise summation, mean aggregation, or channel-wise concatenation before a shared decoder or segmentation head~\cite{hong2021multimodal}. However, accurately modeling complementary features remains impeded by SAR-RGB heterogeneity. Inspired by contrastive learning~\cite{zhao2023cddfuse, wang2025pnas}, recent works align modality-shared features to guide complementary feature learning and alleviate inter-modal conflicts (see \cref{fig1}(b)), yet spatial-domain modeling remains inefficient. 

To better capture RGB-SAR complementarity, some methods explore frequency-domain fusion~\cite{wang2025fcenet, shi2025frefusion, liang2024felnfn}. For example, Li \textit{et al.}~\cite{li2023progressive} treat phase as modality-shared features but discard amplitude information essential for terrain delineation and do not systematically validate phase sharing. 

We conduct a comprehensive spectral analysis (see \cref{fig1}(c)-(g)) to reveal three critical frequency-domain priors: \textit{1)} SAR-RGB spatial-domain differences are anisotropic and susceptible to pixel-level mismatch (see \cref{fig1}(c)); \textit{2)} average phase spectra differ minimally and primarily in low frequencies, indicating phase as shared information (see \cref{fig1}(e)); and \textit{3)} amplitude mismatches exhibit distinct high- and low-frequency patterns, most pronounced in the high-frequency band, highlighting the need for frequency-aware modeling (see \cref{fig1}(d)). 

Motivated by these insights, we propose \textit{Phase-Amplitude Decoupling (PAD)}, which decouples phase components capturing modality-shared structures from amplitude components encoding modality-complementary details. By explicitly aligning shared phase representations, PAD enhances complementary amplitude feature extraction and reduces cross-modal conflicts. PAD comprises two modules: \textit{Phase Spectrum Correction (PSC)}, which aligns and reinforces structural details via convolution-guided scaling; and \textit{Amplitude Spectrum Fusion (ASF)}, which dynamically integrates high- and low-frequency patterns using frequency-adaptive multilayer perceptrons (MLPs). 

Our contributions are fourfold: 

\begin{itemize}

	\item We systematically analyze registered datasets to establish the frequency distributions of SAR-RGB shared (phase) and complementary (amplitude) patterns, providing interpretable physical priors for multi-source fusion. 
	
	\item We introduce frequency-domain amplitude-phase decoupling for multi-modal fusion, orthogonally separating phase (modality-shared) and amplitude (modality-complementary) features to resolve heterogeneity-induced conflicts. 
 
	\item PSC employs convolution-guided multiplicative scaling to align and reinforce critical structural details for robust fusion. 

 	\item ASF leverages frequency-adaptive MLPs to integrate complementary high-frequency (local details) and low-frequency (global structure) components, exploiting SAR’s morphological sensitivity and RGB’s spectral richness for improved segmentation. 

\end{itemize}

\begin{figure*}[!t]
	\centering
	\includegraphics[width = \linewidth]{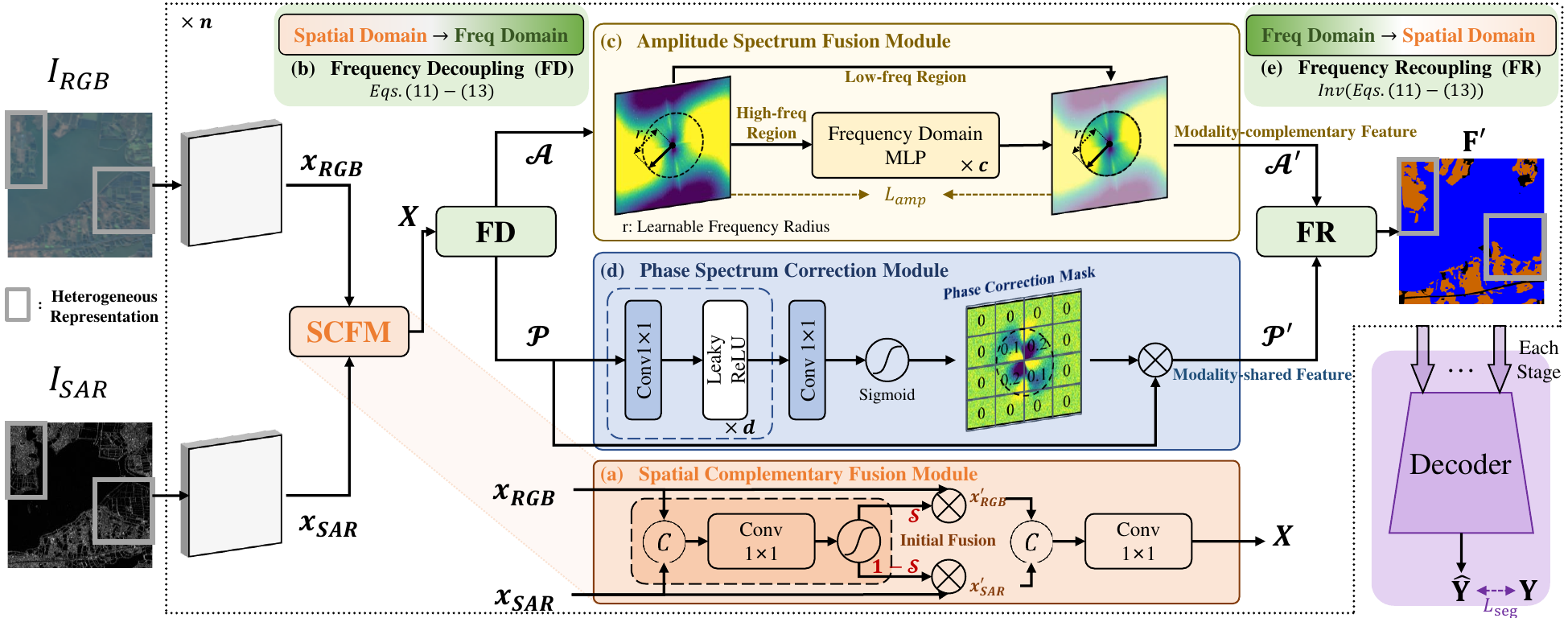}
	\caption{\textbf{Flowchart of the PAD Framework.} Registered RGB and SAR inputs, $I_{\mathrm{RGB}}$ and $I_{\mathrm{SAR}}$, are processed by asymmetric backbones to extract features $x_{\mathrm{RGB}}$ and $x_{\mathrm{SAR}}$ at each stage. These features pass through $n$ PAD fusion modules, each comprising: 
	(a) SCF; 
	(b) FD (spatial-to-frequency transform); 
	(c) ASF; 
	(d) PSC; and 
	(e) FR (frequency-to-spatial transform). 
	At each stage, the fused features are concatenated channel-wise (outer circle, $C$) and fed into a shared decoder. Here, $\mathcal{A}$ and $\mathcal{P}$ denote the amplitude and phase spectra, respectively.}
	\label{fig2}
\end{figure*}

\section{Related Work}

\subsection{Multi-Modal Fusion for Land Cover Classification}

Single-modality approaches~\cite{fu2019danet, yuan2020ocrnet, wu2020cgnet, xie2021segformer, zheng2021SETR, cheng2022masked, xu2023pidnet, chen2021transunet} perform well on homogeneous data but cannot exploit cross-modal synergies in complex environments. SAR-RGB fusion methods generally follow two paradigms: cross-modal transfer learning and hierarchical feature integration.

\subsubsection{Cross-Modal Transfer Learning}

Transfer learning~\cite{li2024assisted} adapts knowledge from one modality to another, but the pronounced differences between SAR and RGB often degrade learned representations, limiting their multi-modal applicability.

\subsubsection{Hierarchical Feature Fusion}

Feature fusion has evolved through three stages: 
\textit{a) Basic Fusion Operators:} Early methods employed concatenation or weighted summation~\cite{hughes2018pscnn}, but these are prone to modality dominance and redundancy. 
\textit{b) Attention-Guided Fusion:} Spatial-channel attention mechanisms alleviate these issues~\cite{Li22mcanet, ren2022ddhr}. For example, MCANet’s cross-modal attention (MCAM) models inter-modality dependencies; MRFS~\cite{zhang2024mrfs} introduces progressive cycle attention for infrared-visible fusion; and CMX~\cite{zhang2023cmx} applies bidirectional channel-spatial attention. However, these paradigms incur high computational costs. 
\textit{c) Complementarity-Aware Fusion:} Recent approaches explicitly decouple shared and complementary features. ASANet~\cite{zhang2024asanet} uses semantic focusing modules to emphasize complementary cues; CDDFuse~\cite{zhao2023cddfuse} applies contrastive learning to separate private and shared representations; and CEN~\cite{wang2020cen} leverages dynamic channel exchange with independent batch normalization. Despite these advances, spatial-domain methods still struggle with geometric misalignment and inefficiency.

\subsection{Frequency Representation Learning in Vision Tasks}

Frequency representation learning has emerged as a powerful tool across various vision tasks due to its capacity for global modeling and structural decomposition. We briefly review three relevant areas:

\subsubsection{Spectral Decomposition for Structure-Style Separation}

Fourier-based decomposition isolates phase (structural semantics) and amplitude (appearance) for domain generalization and style transfer~\cite{zhao2024music}. Early methods relied on feature statistics for style normalization~\cite{huang2017stylenorm, jin2020stylereid}, while recent works apply Fourier transforms for clearer separation, enabling content-preserving augmentation by swapping amplitudes across domains~\cite{chen2021apr, lee2023dac}.

\subsubsection{Frequency-Driven Global Modeling}

Fourier transforms provide a global receptive field useful for efficient attention and context modeling. Fast Fourier Convolution (FFC)~\cite{chi2020ffc} and FFTformer~\cite{kong2023fftformer} leverage spectral processing to replace or accelerate spatial attention. Spectral token mixing has also proven effective in lightweight transformers~\cite{rao2021gfnet, guibas2021afno}.

\subsubsection{Frequency-Based Inductive Biases in Dense Prediction}

Spectral priors benefit dense prediction tasks~\cite{ju2025fourier, xiao2024freq, yu2024phase, cheng2023frequency}. DFF~\cite{lin2023dff} uses frequency-domain dynamic filtering to amplify invariant low-frequency features, improving cross-domain generalization. FreqFusion~\cite{chen2024freqfusion} designs learnable high- and low-pass filters for semantic consistency and boundary refinement. FCENet~\cite{wang2025fcenet} employs frequency-selective modules to model cross-modal spectral correlations in RGB-NIR denoising.

In summary, frequency-based representations offer a flexible and physically interpretable design space for vision tasks. Yet, few frameworks address task-specific fusion in remote sensing. We target this gap in the next section.

\subsection{Frequency-Domain Fusion Methods}

The application of frequency representations to multi-modal fusion is relatively recent. Shi \textit{et al.}~\cite{shi2025frefusion} introduce Frefusion, which applies spectral attention to merge modalities in the frequency domain. Frequency spectra naturally decompose signals into phase (modality-shared structures) and amplitude (modality-complementary details), offering a principled basis for fusion.

However, existing methods often transplant spatial operators, such as MLPs~\cite{liang2024felnfn}, convolutions~\cite{wang2025fcenet}, or attention~\cite{shi2025frefusion}, into the spectral domain, modulating magnitude while neglecting the distinct roles of phase and amplitude. As a result, shared information may be preserved, but complementary cues are frequently suppressed.

In contrast, our PAD framework explicitly decouples phase and amplitude spectra. Through systematic spectral analysis (see \S\ref{sec:iii_a}), we show that SAR and RGB exhibit strong phase consistency, while amplitude captures complementary patterns. PAD introduces 
\textit{1) a Phase Spectrum Correction (PSC) module}, which performs robust structural alignment, and 
\textit{2) an Amplitude Spectrum Fusion (ASF) module}, which conducts frequency-adaptive refinement. 
This decomposition-correction-fusion pipeline distinguishes PAD from previous FFT-based approaches by providing a physically interpretable strategy for multi-modal fusion.

The two most related works, FeINFN~\cite{liang2024felnfn} and SFINet++~\cite{zhou2024SFINetplusplus}, also leverage amplitude and phase in the frequency domain for high-frequency detail enhancement and global context modeling. However, they lack a physically grounded interpretation of amplitude-phase roles and do not explicitly decouple shared versus complementary information. In contrast, PAD is the first framework to formulate and exploit amplitude-phase decomposition for principled alignment and complementary integration across modalities.

\section{Proposed Method}

To address the fundamental conflict between modality-shared geometry and modality-complementary appearance in SAR-RGB fusion, we introduce \textit{Phase-Amplitude Decoupling (PAD)}, a frequency-aware multi-modal fusion framework. The overall architecture in \cref{fig2} comprises three modules: the \textbf{Spatial Complementary Fusion (SCF)} module, which preserves fine-grained spatial priors; the \textbf{Phase Spectrum Correction (PSC)} module, which aligns and reinforces shared structural details; and the \textbf{Amplitude Spectrum Fusion (ASF)} module, which adaptively integrates complementary high- and low-frequency amplitude cues. Unlike previous methods that apply spectral operations uniformly, PAD explicitly decouples and then recombines phase and amplitude with dedicated modules for physics-consistent fusion. 

\begin{algorithm}
\caption{\small \textbf{Spectral Difference Analysis.}}
\footnotesize
\label{alg:spec_diff}
\SetAlgoLined
\DontPrintSemicolon
\SetKwInOut{Input}{Input}\SetKwInOut{Output}{Output}
\SetKwComment{Comment}{\footnotesize//}{{\footnotesize}} 
\SetKwComment{StepOne}{\textcolor{black}{\sffamily\bfseries 1: Preprocessing}}{} 
\SetKwComment{StepTwo}{\textcolor{black}{\sffamily\bfseries 2: Spectral Decomposition}}{} 
\SetKwComment{StepThree}{\textcolor{black}{\sffamily\bfseries 3: Dataset-level Aggregation}}{} 
\Input{SAR/RGB image pairs $\{I_{\mathrm{SAR}}^{(i)}, I_{\mathrm{RGB}}^{(i)}\}_{i = 1}^N$}
\Output{RSD, RAD, APPD metrics for each image pair}
\BlankLine

\StepOne{}
	\Comment{Convert RGB image to grayscale}
	$I_{\mathrm{RGB}} \gets 0.299 R + 0.587 G + 0.114 B$ 
	\Comment{Retain SAR image as is}
	$I_{\mathrm{SAR}} \gets I_{\mathrm{SAR}}$ 

\StepTwo{}
 \For{each pair $(i)$}{
	\Comment{Frequency domain conversion}
	Compute $\mathcal{A}_{\mathrm{SAR}}^{(i)}$ and $\mathcal{P}_{\mathrm{SAR}}^{(i)}$ via 2D FFT and FFTshift; 
	Compute $\mathcal{A}_{\mathrm{RGB}}^{(i)}$ and $\mathcal{P}_{\mathrm{RGB}}^{(i)}$ similarly 
	
	\Comment{Spatial difference computation}
	Compute $\Delta I^{(i)} \gets \mathrm{RSD}^{(i)}$ via \cref{eq:rsd} 
	\Comment{Amplitude difference computation}
	Compute $\Delta \mathcal{A}^{(i)} \gets \mathrm{RAD}^{(i)}$ via \cref{eq:rad} 
	\Comment{Phase difference computation}
	Compute $\Delta \mathcal{P}^{(i)} \gets \mathrm{APPD}^{(i)}$ via \cref{eq:appd} 
}

\StepThree{}
	\Comment{Mean spatial, amplitude, phase difference computation}
 $\mathrm{RSD} \gets \mathtt{Mean}(\{\Delta I^{(i)}\}_{i = 1}^N)$ 
 $\mathrm{RAD} \gets \mathtt{Mean}(\{\Delta \mathcal{A}^{(i)}\}_{i = 1}^N)$ 
 $\mathrm{APPD} \gets \mathtt{Mean}(\{\Delta \mathcal{P}^{(i)}\}_{i = 1}^N)$ 
\end{algorithm}

\subsection{Frequency-Domain Priors for Amplitude-Phase Decoupling} \label{sec:iii_a}

To validate the physical significance of amplitude-phase decoupling and reveal the shared and complementary relations between RGB and SAR, we conduct a systematic spectral analysis on \textsc{WHU-OPT-SAR}, the largest RGB-SAR segmentation benchmark.

First, we visualize averaged \textit{difference spectra} (see \cref{fig1}(c)-(e)) using \cref{alg:spec_diff} to compare modality differences in the spatial, amplitude, and phase domains. This analysis shows that frequency-domain representations capture cross-modal sharing and complementarity more effectively than spatial representations.

The difference spectra are defined as: 
\begin{align}
	\mathrm{RSD} & = \frac{1}{N} \sum_{i = 1}^N \frac{\left| I_{\mathrm{SAR}}^{(i)} - I_{\mathrm{RGB}}^{(i)} \right|}{I_{\mathrm{RGB}}^{(i)} + \epsilon}, \label{eq:rsd} \\
	\mathrm{RAD} & = \frac{1}{N} \sum_{i = 1}^N \frac{\left| \mathcal{A}_{\mathrm{SAR}}^{(i)} - \mathcal{A}_{\mathrm{RGB}}^{(i)} \right|}{\mathcal{A}_{\mathrm{RGB}}^{(i)} + \epsilon}, \label{eq:rad} \\
	\mathrm{APPD} & = \frac{1}{N} \sum_{i = 1}^N \left| \arg \exp \left(\mathrm{i} (\mathcal{P}_{\mathrm{SAR}}^{(i)} - \mathcal{P}_{\mathrm{RGB}}^{(i)}) \right) \right|, \label{eq:appd}
\end{align}
where $I$ denotes pixel intensities; $\mathcal{A}$ and $\mathcal{P}$ are amplitude and phase spectra, respectively; $\arg(\cdot)$ returns the complex angle; $\mathrm{i} = \sqrt{-1}$; and $\epsilon > 0$ prevents division by zero. By leveraging the periodicity of complex exponentials, \cref{eq:appd} maps phase differences to $[-\pi,\pi)$, correcting wrap-around artifacts (\textit{e.g.}, $2\pi \to 0$).

\Cref{fig1}(c)-(e) reveals three priors: 
\textit{1)} spatial differences (RSD) exhibit anisotropic distributions; 
\textit{2)} amplitude differences (RAD) concentrate in high-frequency regions; and 
\textit{3)} phase differences (APPD) remain minimal, with slight low-frequency clustering. 
These frequency-domain metrics provide better explainability than spatial RSD, confirming that spectral decomposition more faithfully captures cross-modal characteristics.

Physically, low-frequency phase disparities stem from geometric distortions (\textit{e.g.}, side-looking SAR \textit{vs.} orthographic RGB), while high-frequency variations arise from SAR speckle. Amplitude disparities reflect modality-complementary sensitivity to object scales: low frequencies encode large structures (farmlands, forests), whereas high frequencies capture fine details (roads).

These analyses yield two key insights that guide our design: 
\textbf{Phase sharing.} Consistently low APPD indicates strong phase consistency; PSC therefore performs pixel-level alignment while preserving this shared information. 
\textbf{Amplitude complementarity.} The frequency-dependent divergence of RAD motivates ASF’s dynamic frequency-radius adjustment to handle amplitude variations.

Second, we conduct sensitivity experiments on sampling rates (see \cref{fig1}(f)-(g)). By downsampling images at various ratios and computing the mean/variance of RSD, RAD, and APPD across the dataset, we observe that RAD and APPD remain nearly constant, whereas RSD fluctuates more. This invariance indicates that frequency-domain complementarity arises from intrinsic modality characteristics rather than sampling artifacts. 

\begin{table}[!t]
	\centering
	\caption{\textbf{Statistical Analysis of the Average Periodic Phase Difference (APPD) Spectrum.}}
	\label{table1}
	\begin{tabular}{l|ccc}
	\toprule[1.1pt]
	Metric & \makecell{Low frequency \\ (LF)} & \makecell{High frequency \\ (HF)} & ALL \\
	\midrule
	Shapiro-Wilk statistic & 1.0044 & 1.0050 & 1.0131 \\
	Shapiro-Wilk p-value & 1.0000 & 1.0000 & 1.0000 \\
	Skewness & -0.0521 & 0.0020 & -0.0265 \\
	Kurtosis & 0.0750 & -0.0117 & 0.0322 \\
	\bottomrule[1.1pt]
	\end{tabular}
\end{table}

\begin{figure}[!t]
	\centering
	\includegraphics[width = \linewidth]{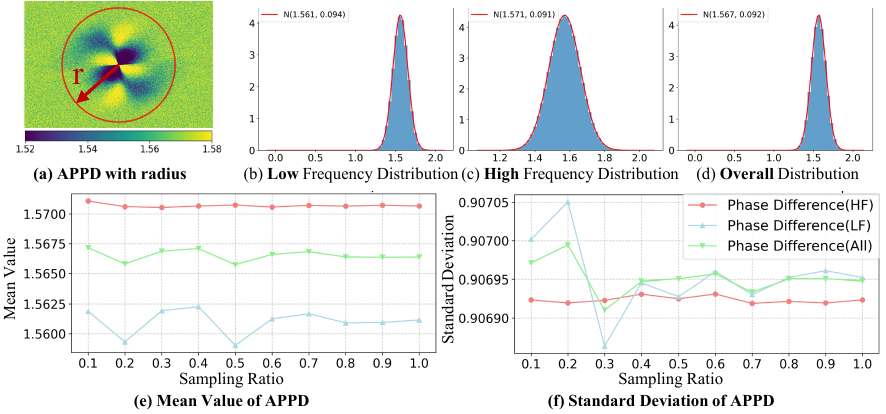}
	\caption{\textbf{Statistical Analysis of APPD.} 
	(a) APPD spectrum. 
	(b-d) Frequency histograms with Gaussian fits for low frequency (LF; radius = $0.5\times$ half-diagonal), high frequency (HF), and all spectra (ALL). 
	(e-f) Mean-variance trajectories across sampling rates. 
	Across sampling rates, all spectra exhibit near-Gaussian behavior, while slight deviations in LF and ALL are attributable to structural noise at low frequencies, indicating that the phase, particularly HF, serves as a robust shared feature.}
	\label{fig3}
\end{figure}

Third, we assess statistical consistency of cross-modal phase spectra via Shapiro-Wilk tests and by computing skewness and kurtosis of APPD within each frequency band (see \cref{table1}). Frequency histograms and mean-variance trajectories (see \cref{fig3}) show that APPD distributions closely follow a Gaussian profile, especially in high-frequency regions; small low-frequency deviations likely result from structural noise. In summary, RGB and SAR phases are highly similar and largely unaffected by instance-level variations. 

\begin{figure}[!t]
	\centering
	\includegraphics[width = \linewidth]{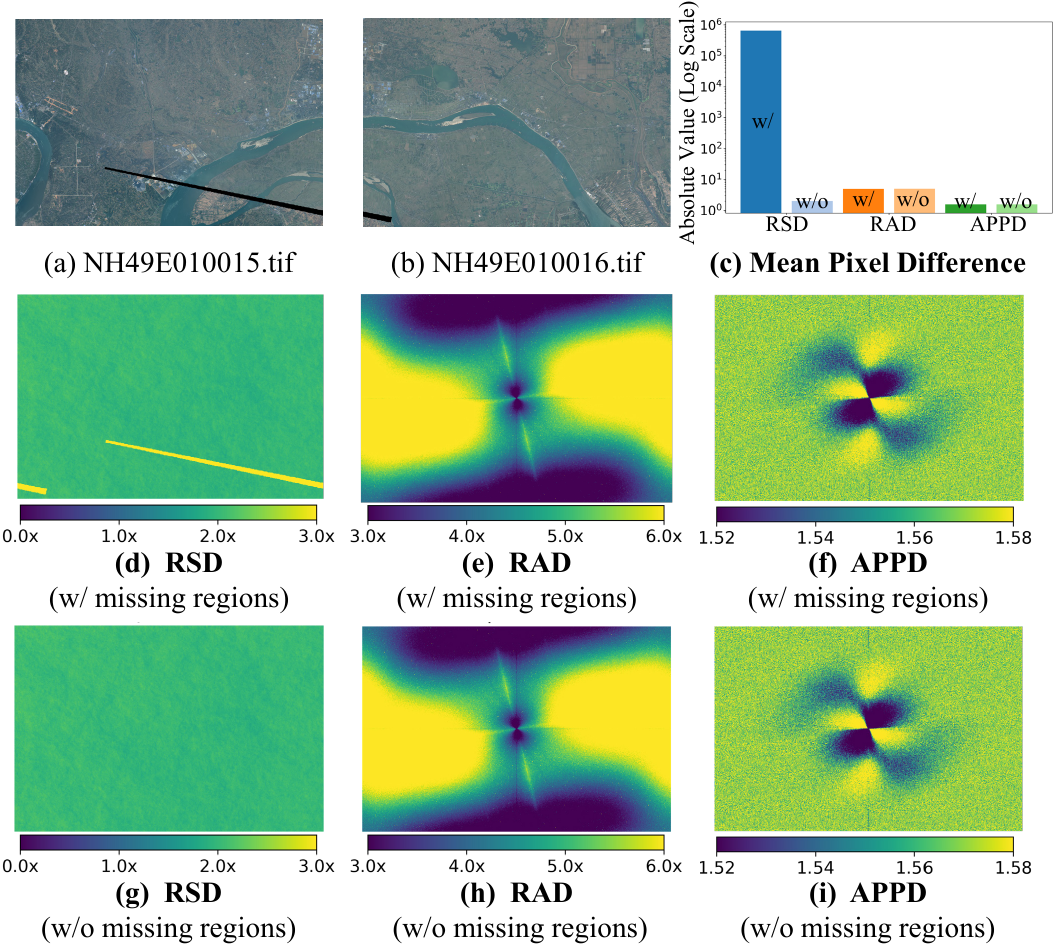}
	\caption{\textbf{Visualization of Spatial-Frequency Spectrum Differences under Missing RGB Regions.} 
	(a-b) Two RGB images have missing regions in the original dataset.
	(c) Variation in RSD, RAD, and APPD, indicating RSD’s greater sensitivity to missing data. 
	(d-f) Difference spectra computed \textit{including} the missing regions. 
	(g-i) Difference spectra computed \textit{excluding} the missing regions. 
	Overall, frequency-domain metrics (RAD and APPD) are more robust to missing data than the spatial-domain RSD.}
	\label{fig4}
\end{figure}

\begin{figure}[!t]
	\centering
	\includegraphics[width = \linewidth]{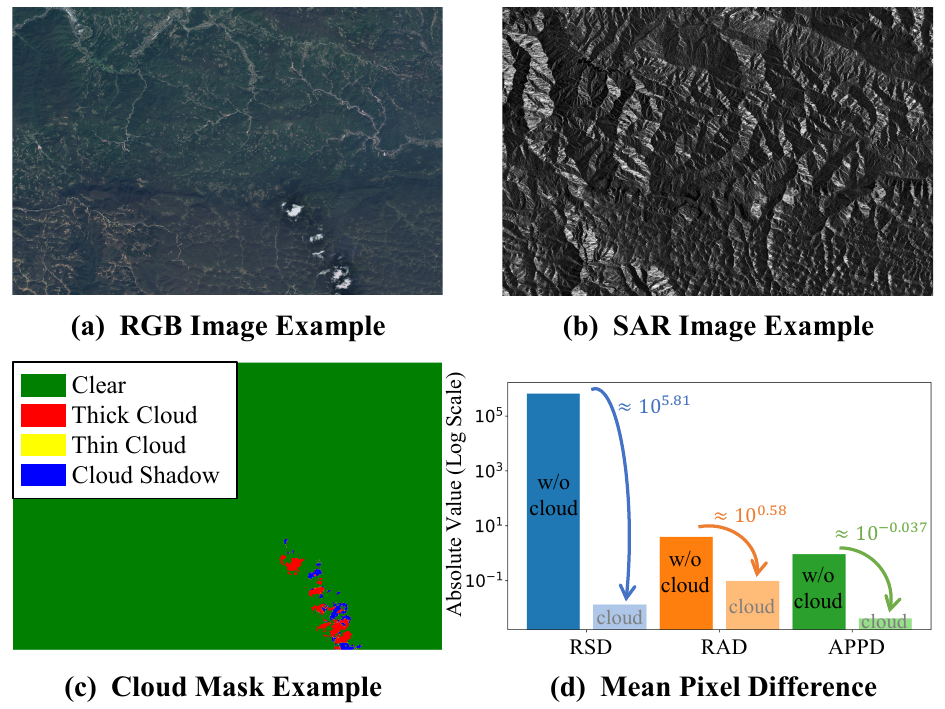}
	\caption{\textbf{Visualization of Spatial-Frequency Spectrum Differences under Cloud Obscuration.} 
	(a-c) Example observations with cloud masks derived from the RGB, SAR, and NIR channels in \textsc{WHU-OPT-SAR}~\cite{wright2025cloud}, respectively. 
	(d) Comparison of RSD, RAD, and APPD between cloud-covered and cloud-free areas; arrows indicate the value of the decrease.}
	\label{fig5}
\end{figure}

Finally, we evaluate phase invariance under two challenging scenarios: 
\textit{1)} partial RGB removal (see \cref{fig4}), and 
\textit{2)} cloud occlusion (see \cref{fig5}). 
In the former, RSD increases markedly when RGB pixels are missing (see \cref{fig4}(d) and (g)), whereas APPD remains stable (see \cref{fig4}(f) and (i)). In the latter, the variance of APPD ($\Delta \approx 10^{-0.037}$) is negligible compared with that of RSD ($\Delta \approx 10^{5.81}$) and RAD ($\Delta \approx 10^{0.58}$) in \cref{fig5}(d). These results confirm that phase, particularly at high frequencies, retains robust cross-modal consistency, underscoring its role as a shared feature to mitigate inter-modal conflicts and enhance modality-complementary fusion.

\subsection{Overall Framework}

As illustrated in \cref{fig2}, PAD integrates SAR and RGB via a dual-branch encoder and a shared decoder, with its primary innovation lying in cross-modal feature fusion through phase-amplitude decoupling. The encoder employs \textit{asymmetric convolutional backbones} $\mathcal{B}_{\mathrm{SAR}}$ and $\mathcal{B}_{\mathrm{RGB}}$, each with independent parameters to preserve modality-complementary attributes. Specifically, $\mathcal{B}_{\mathrm{SAR}}$ extracts microwave backscattering features from SAR images, while $\mathcal{B}_{\mathrm{RGB}}$ captures optical reflectance patterns from RGB images. Both backbones produce hierarchical features at four stages with resolutions $\{1/4,1/8,1/16,1/32\}$ for $i\in\{0,1,2,3\}$:
\begin{align}
	x_{\mathrm{RGB}}^{i} = \mathcal{B}_{\mathrm{RGB}}^{(i)} \left(I_{\mathrm{RGB}} \right), \quad
	x_{\mathrm{SAR}}^{i} = \mathcal{B}_{\mathrm{SAR}}^{(i)} \left(I_{\mathrm{SAR}} \right).
\end{align}

Although frequency-domain fusion is robust to modality differences, it may discard rich local interactions. To retain fine-grained spatial priors, we introduce a lightweight SCF module. SCF computes pixel-wise fusion weights via spatial attention gating and reduces feature dimensionality to constrain the solution space for subsequent spectral operations:
\begin{subequations}\label{eq:SCF}
\begin{align}
	X^{i} & = \mathrm{SCF} \left(x_{\mathrm{RGB}}^{i}, x_{\mathrm{SAR}}^{i} \right), \label{eq:SCF_overall} \\
	\mathcal{S} & = \sigma \left(\mathrm{Conv}_{1\times1} \left(\mathrm{Concat} \left(x_1,x_2 \right) \right) \right), \label{eq:SCF_att} \\
	x_1' & = \mathcal{S}\odot x_1, \quad 
	x_2' = \left(1-\mathcal{S} \right)\odot x_2, \label{eq:SCF_select} \\
	x & = \mathrm{Conv}_{1\times1} \left(\mathrm{Concat} \left(x_1',x_2' \right) \right), \label{eq:SCF_fuse}
\end{align}
\end{subequations}
where $(x_1,x_2) = (x_{\mathrm{RGB}}^{i},x_{\mathrm{SAR}}^{i})$; $\sigma(\cdot)$ is the Sigmoid; $\mathrm{Conv}_{1\times1}$ denotes a $1\times1$ convolution; $\mathrm{Concat}(\cdot,\cdot)$ concatenates along channels; and $\odot$ denotes the Hadamard (element-wise) product.

The fused feature $X^{i}$ is then transformed to separate amplitude $\mathcal{A}^{i}$ and phase $\mathcal{P}^{i}$ spectra:
\begin{align}
	\mathcal{A}^{i}, \mathcal{P}^{i} = \mathrm{FD} \left(X^{i}\right),
\end{align}
where $\mathrm{FD}(\cdot)$ denotes the frequency decoupling operator (defined in \S\ref{sec:freq_decouple}). ASF enhances the amplitude spectrum using a trainable frequency radius $r$ to distinguish and reinforce high-frequency details while preserving low-frequency structures:
\begin{align}
	\mathcal{A}'^{i} = \mathrm{ASF} \left(\mathcal{A}^{i}; r \right).
\end{align}
PSC applies multiplicative scaling to correct the phase:
\begin{align}
	\mathcal{P}'^{i} = \mathrm{PSC} \left(\mathcal{P}^{i} \right).
\end{align}
Finally, inverse reconstruction produces fused features:
\begin{align}
	\bm{F}'^{i} = \mathrm{FR} \left(\mathcal{A}'^{i}, \mathcal{P}'^{i} \right),
\end{align}
where $\mathrm{FR}(\cdot,\cdot)$ is the inverse frequency recoupling operator (defined in \S\ref{sec:freq_decouple}). A shared decoder $\mathcal{D}$ aggregates multi-stage fused features to generate the land-cover map:
\begin{align}
	\widehat{\bm{Y}} = \mathcal{D} \left(\bm{F}'^{0}, \bm{F}'^{1}, \bm{F}'^{2}, \bm{F}'^{3} \right).
\end{align}

\subsection{Frequency Decoupling and Recoupling} \label{sec:freq_decouple}

To obtain physically decoupled spectral representations, we employ a bidirectional transformation comprising frequency decoupling (FD) and inverse recoupling (FR). Given $X\in\mathbb{R}^{H\times W}$, a 2D real-input fast Fourier transform (rFFT)~\cite{cooley1965fft} yields:
\begin{align}
	\mathcal{F}(u,v) = \sum_{y = 0}^{H-1}\sum_{x = 0}^{W-1} X(y,x) 
	\exp \left(-\mathrm{i}2\pi \left(\frac{ux}{W} + \frac{vy}{H} \right) \right),
\label{eq:fft}
\end{align}
where $(u,v)$ are frequency coordinates. Exploiting conjugate symmetry for real inputs, we retain only the non-redundant region and center low-frequency components via an FFT shift:
\begin{align}
	\mathcal{F}_{\mathrm{shift}}(u,v) = \mathcal{F} \left(u \oplus \frac{W}{2}, v \oplus \frac{H}{2} \right),
\label{eq:fft_shift}
\end{align}
where $\oplus$ is modulo addition. Amplitude and phase are:
\begin{align}
	\mathcal{A} = \left|\mathcal{F}_{\mathrm{shift}} \right|, \quad
	\mathcal{P} = \angle \mathcal{F}_{\mathrm{shift}},
\label{eq:fft_decoupling}
\end{align}
with $\mathcal{A},\mathcal{P}\in\mathbb{R}^{H\times(\lfloor W/2\rfloor+1)}$. FR applies the inverse operations to reconstruct spatial features from $(\mathcal{A},\mathcal{P})$.


\subsection{Phase Spectrum Correction (PSC) Module}


While prior works often assume perfect phase alignment~\cite{li2023progressive}, our spectral analysis (see \cref{fig1}(e)) shows that low-frequency geometric inconsistencies persist, necessitating a learnable, spatially aware correction mechanism. Inspired by channel attention, we propose a lightweight \textit{Phase Spectrum Correction} (PSC) module tailored for frequency-domain phase alignment.

As shown in \cref{fig1}(e), minor misalignments in the low-frequency phase spectra between SAR and RGB, attributable to geometric distortions (\textit{e.g.}, side-looking SAR \textit{vs.} orthographic RGB), can compromise spatial consistency and boundary precision. Traditional rigid registration is inadequate for such non-rigid distortions. PSC performs adaptive geometric correction using learnable multiplicative masks.

Unlike global channel scaling in traditional attention, PSC estimates a spatially varying modulation mask via a stack of $d$ consecutive $1{\times}1$ convolutions, enabling localized, fine-grained phase adjustment. Given an input phase-spectrum tensor $P \in \mathbb{R}^{C\times H\times (\lfloor W/2 \rfloor + 1)}$, we first compress channels with $d$ successive $1{\times}1$ convolutions (reducing $C$ to $C/r$, $r \in \{4,8\}$), each followed by a Leaky ReLU~\cite{maas2013leakyrelu}. A final $1{\times}1$ convolution produces a single-channel \textit{Phase Correction Mask} (PCM), constrained to $(0,1)$ by a Sigmoid:
\begin{align}
	\mathrm{PCM} = \sigma \left(\mathrm{Conv}^{1\times1}_{d} \left(\delta \left(\cdots \delta \left(\mathrm{Conv}^{1\times1}_{1} \left(P \right) \right) \cdots \right) \right) \right),
\end{align}
where $\delta(\cdot)$ is Leaky ReLU and $\mathrm{Conv}^{1\times1}_{k}$ denotes the $k$-th $1{\times}1$ convolution in the stack. Pixel-wise multiplicative scaling is then applied:
\begin{align}
	P' = P \odot \left(1 + \mathrm{PCM} \right).
\end{align}
This formulation enables localized, fine-grained phase adjustment while preserving phase periodicity and incurring minimal computational overhead.

\subsection{Amplitude Spectrum Fusion (ASF) Module}

Previous frequency-domain methods often employ fixed high-/low-pass filters or spatially uniform spectral attention, which cannot adapt to content-specific frequency distributions. In contrast, our analysis in \cref{fig1}(d) shows that RGB-SAR spectral discrepancies are concentrated in medium-to-high amplitude frequencies. {To exploit this prior while maintaining gradient stability, ASF uses a \textit{trainable} frequency radius to dynamically separate and enhance complementary bands via three steps.} 
First, for each spectral component indexed by $i$, we compute a normalized radial distance:
\begin{align}
	\mathrm{Dist}(i) = \frac{\sqrt{\left(X_i - X_{\mathrm{c}} \right)^2 + \left(Y_i - Y_{\mathrm{c}} \right)^2}}{\sqrt{X_{\max}^{2} + Y_{\max}^{2}}},
\end{align}
where $(X_{\mathrm{c}},Y_{\mathrm{c}})$ is the spectrum center and $(X_{\max},Y_{\max})$ are the maximal coordinate indices along each axis, so $\mathrm{Dist}(i) \in [0,1]$ quantifies each component’s position relative to the center. 
A trainable boundary radius $r$ is mapped by a Sigmoid to $R = \sigma(r) \in [0,1]$, partitioning the spectrum into low-/high-frequency regions and preventing divergence during training. To maintain differentiability, we use a soft mask:
\begin{align}
	\mathrm{Mask}_{\mathrm{hi}} = \sigma \left(\left(\mathrm{Dist}-R \right)\cdot \tau \right),
\end{align}
where $\tau = 10$ is a temperature hyperparameter. High-frequency components are enhanced via a position-wise MLP:
\begin{align}
	A_{\mathrm{hi}} = \mathrm{MLP} \left(A \odot \mathrm{Mask}_{\mathrm{hi}}\right),
\end{align}
and then concatenated with the original amplitude $A$ and fused by a $1{\times}1$ convolution with a residual update:
\begin{align}
	A' = \mathrm{Conv}_{1\times1} \left(\mathrm{Concat} \left(A, A_{\mathrm{hi}} \right) \right), \quad
	A \leftarrow A + A',
\end{align}
where $A \in \mathbb{R}^{C\times H\times (\lfloor W/2 \rfloor + 1)}$ is the input amplitude spectrum, and $\mathrm{MLP}(\cdot)$ is a channel-wise two-layer perceptron with nonlinearity. This preserves low-frequency structure while adaptively reinforcing high-frequency details carrying modality-complementary semantics. The learnable radius $R$ self-adjusts to input characteristics, eliminating manual threshold tuning.

\subsection{Loss Function}

We integrate three objectives to jointly promote semantic accuracy, multi-scale representation learning, and amplitude-phase coherence.

\subsubsection{Primary Semantic Supervision}

\begin{align}
	\mathcal{L}_{\mathrm{seg}} = -\sum_{i \in \Omega}\sum_{c = 1}^{C} y_{i,c} \log \hat{p}_{i,c},
\end{align}
where $\Omega$ is the set of labeled pixels (excluding the ignore label 255), $y_{i,c}\in\{0,1\}$ is the one-hot ground truth for class $c$, and $\hat{p}_{i,c}$ is the predicted probability.

\subsubsection{Auxiliary Semantic Supervision}

An auxiliary head at the penultimate stage ($t = 2$) uses the same objective:
\begin{align}
	\mathcal{L}_{\mathrm{aux}} = -\sum_{i \in \Omega}\sum_{c = 1}^{C} y_{i,c} \log \hat{p}^{(a)}_{i,c},
\end{align}
where $\hat{p}^{(a)}_{i,c}$ denotes auxiliary predictions.

\subsubsection{Phase-Amplitude Decoupling Constraint}

To regularize the fused amplitude towards its decoupled counterpart across scales, we define:
\begin{align}
	\mathcal{L}_{\mathrm{amp}} = \frac{1}{\left|\mathcal{T} \right|}\sum_{t \in \mathcal{T}} \alpha_t \left\| \mathcal{A}'_{t} - \mathcal{A}_{t} \right\|_{F}^{2},
\end{align}
where $\mathcal{T} = \{0,1,2,3\}$ indexes feature scales; $\alpha_t = 0.5 + 0.5 t/3$ applies depth-progressive weighting; $\|\cdot\|_{F}$ is the Frobenius norm; and $\mathcal{A}_{t}, \mathcal{A}'_{t}$ are the original and fused amplitude spectra at scale $t$.

\subsubsection{Composite Objective}

\begin{align}
	\mathcal{L}_{\mathrm{total}} = \mathcal{L}_{\mathrm{seg}} + \lambda_{1} \mathcal{L}_{\mathrm{aux}} + \lambda_{2} \mathcal{L}_{\mathrm{amp}},
\end{align}
where $\lambda_{1} = 0.4$ balances auxiliary supervision and $\lambda_{2} = 0.1$ controls amplitude regularization. This composite loss promotes geometric invariance by stabilizing phase-sensitive features via $\mathcal{L}_{\mathrm{amp}}$, and semantic robustness via hierarchical amplitude refinement.

\section{Experimental Results}

\subsection{Experimental Setup}

\subsubsection{Dataset Description}

\textsc{WHU-OPT-SAR}~\cite{Li22mcanet} is a multi-modal land-use dataset covering Hubei Province, China. It contains 100 co-registered RGB-SAR-NIR triples labeled with seven categories: farmland, urban, village, water, forest, roads, and others. RGB (GF-1) and SAR (GF-3) images are resampled to 5 m resolution and sized $5556 \times 3704$ pixels. Each image is cropped into non-overlapping $512 \times 512$ patches, yielding 7{,}000 samples split 80\%/20\% for training/testing.

\textsc{DDHR-SK}~\cite{ren2022ddhr} is a fog-degraded subset of DDHR from Pohang, South Korea, pairing GF-2 RGB and GF-3 SAR images. It contains 6{,}173 cloud-affected RGB-SAR pairs, resampled to 1 m and resized to $256 \times 256$ pixels, with five categories: buildings, farmland, greenery, water, and roads. The public split provides 3{,}087 training and 3{,}086 validation samples.

\subsubsection{Training Settings}

All models are trained using the MMSegmentation framework~\cite{mmseg2020} on a single NVIDIA H800 GPU (comparisons use an NVIDIA A800). We use a batch size of 4 for 80k iterations, validating every 8k iterations. Optimization is performed with AdamW, an initial learning rate of $1\times10^{-4}$, and a cosine-annealing schedule (T\_max = total iterations). A fixed seed (42) ensures reproducibility.

\subsubsection{Evaluation Metrics}

We report mean Intersection-over-Union (mIoU), overall accuracy (OA), F1 score, and Cohen’s Kappa (mKappa), selecting the checkpoint with the highest mIoU. To mitigate class-imbalance bias, Kappa is computed for each test sample and then averaged across all samples to obtain mKappa.

\subsection{Main Quantitative Results}
\label{sec:iv_c}

To assess PAD’s effectiveness, we replicate the original configurations of 20 comparison models, 12 state-of-the-art multi-modal segmentation approaches and 8 leading unimodal methods, on both \textsc{WHU-OPT-SAR} and \textsc{DDHR-SK}. 

\begin{table*}[!t]
	\centering
	\caption{\textbf{Quantitative Results (\%) on \textsc{WHU-OPT-SAR}.} Per-class IoU is reported for seven land-cover categories (Farmland, City, Village, Water, Forest, Road, and Others), along with OA, mKappa, mF1, and mIoU. The best and second-best results are highlighted in \textbf{bold} and \underline{underlined}, respectively.}
	\label{table2}
	\begin{tabular}{c|ll|ccccccc|cccc}
	\toprule[1.1pt]
	& \multirow{2}[2]{*}{Method} & \multirow{2}[2]{*}{Venue} & \multicolumn{7}{c|}{Per-class IoU} & \multirow{2}[2]{*}{OA} & \multirow{2}[2]{*}{mKappa} & \multirow{2}[2]{*}{mF1} & \multirow{2}[2]{*}{mIoU} \\
	\cmidrule(lr){4-10}
	& & & Farmland & City & Village & Water & Forest & Road & Others & & & & \\
	\midrule
	\multirow{8}{*}{\rotatebox{90}{RGB}}
	& DANet~\cite{fu2019danet} & CVPR’19 & 67.51 & 55.22 & 46.71 & 60.85 & 83.13 & 38.89 & 20.43 & 83.19 & 49.74 & 67.40 & 53.25 \\
	& CGNet~\cite{wu2020cgnet} & TIP’20 & 61.33 & 45.96 & 37.98 & 49.33 & 80.98 & 1.92 & 0.19 & 79.26 & 37.42 & 50.54 & 39.67 \\
	& OCRNet~\cite{yuan2020ocrnet} & ECCV’20 & 67.68 & 56.48 & 47.47 & 61.22 & 82.98 & 37.20 & 17.97 & 83.23 & 49.11 & 66.95 & 53.00 \\
	& SETR~\cite{zheng2021SETR} & CVPR’21 & 65.25 & 55.07 & 42.97 & 57.80 & 82.44 & 31.91 & 18.36 & 81.94 & 46.19 & 64.73 & 50.54 \\
	& SegFormer~\cite{xie2021segformer} & NeurIPS’21 & 68.18 & 56.91 & 46.82 & 60.85 & 83.19 & 35.24 & 18.73 & 83.37 & 49.46 & 66.79 & 52.84 \\
	& Mask2Former~\cite{cheng2022masked} & CVPR’22 & 67.09 & 56.64 & 43.97 & 60.09 & 81.47 & 36.24 & 21.03 & 82.69 & 47.69 & 66.65 & 52.36 \\
	& PIDNet~\cite{xu2023pidnet} & CVPR’23 & 66.20 & 55.20 & 44.95 & 58.90 & 82.20 & 36.36 & 16.40 & 82.25 & 47.20 & 65.53 & 51.46 \\
	& TransUNet~\cite{chen2021transunet} & MIA’24 & 67.62 & 56.28 & 46.08 & 60.66 & 83.25 & 38.76 & 19.32 & 83.18 & 49.89 & 67.20 & 53.14 \\
	\midrule
	\multirow{8}{*}{\rotatebox{90}{SAR}}
	& DANet~\cite{fu2019danet} & CVPR’19 & 60.42 & 54.12 & 33.05 & 54.45 & 80.57 & 24.89 & 15.11 & 79.44 & 35.28 & 60.16 & 46.09 \\
	& CGNet~\cite{wu2020cgnet} & TIP’20 & 56.72 & 46.81 & 26.32 & 46.30 & 77.53 & 0.27 & 0.00 & 76.39 & 24.64 & 47.00 & 36.28 \\
	& OCRNet~\cite{yuan2020ocrnet} & ECCV’20 & 59.81 & 52.48 & 31.21 & 54.03 & 79.57 & 22.49 & 13.92 & 78.94 & 32.62 & 58.74 & 44.79 \\
	& SETR~\cite{zheng2021SETR} & CVPR’21 & 60.35 & 52.29 & 31.41 & 53.56 & 80.15 & 18.66 & 15.18 & 79.12 & 34.25 & 58.33 & 44.51 \\
	& SegFormer~\cite{xie2021segformer} & NeurIPS’21 & 60.17 & 54.71 & 32.30 & 54.07 & 79.79 & 21.02 & 15.13 & 79.31 & 33.60 & 59.24 & 45.31 \\
	& Mask2Former~\cite{cheng2022masked} & CVPR’22 & 60.39 & 52.86 & 35.69 & 53.85 & 78.36 & 29.16 & 17.20 & 79.26 & 32.76 & 61.35 & 46.79 \\
	& PIDNet~\cite{xu2023pidnet} & CVPR’23 & 58.59 & 52.35 & 29.65 & 53.60 & 79.35 & 23.06 & 13.74 & 78.42 & 32.15 & 58.33 & 44.34 \\
	& TransUNet~\cite{chen2021transunet} & MIA’24 & 61.93 & 54.85 & 33.42 & 56.26 & 80.63 & 27.56 & 14.96 & 80.14 & 37.45 & 61.13 & 47.09 \\
	\midrule
	\multirow{13}{*}{\rotatebox{90}{SAR \& RGB}}
	& MCANet~\cite{Li22mcanet} & JAG’21 & 65.14 & 53.97 & 45.43 & 54.59 & 81.87 & 31.37 & 13.43 & 81.74 & 44.28 & 63.37 & 49.40 \\
	& DDHRNet~\cite{ren2022ddhr} & JAG’22 & 64.80 & 49.42 & 40.98 & 54.11 & 81.31 & 22.96 & 10.32 & 80.94 & 44.49 & 59.84 & 46.27 \\
	& CEN~\cite{wang2020cen} & TPAMI’22 & 67.60 & 55.73 & 46.46 & 60.65 & 83.28 & 34.22 & 15.94 & 83.17 & 49.37 & 65.79 & 51.98 \\
	& TokenFusion~\cite{wang2022tokenfusion} & CVPR’22 & 68.43 & 58.34 & 45.76 & 61.09 & 83.79 & 32.03 & 15.20 & 83.65 & 49.45 & 65.67 & 52.09 \\
	& CDDFuse~\cite{zhao2023cddfuse} & CVPR’23 & 60.07 & 48.64 & 23.96 & 50.67 & 79.47 & 15.16 & 11.43 & 78.02 & 51.36 & 54.54 & 41.34 \\
	& CMX~\cite{zhang2023cmx} & TITS’23 & \textbf{70.04} & \textbf{58.41} & 49.04 & \underline{64.77} & \underline{84.12} & \underline{41.76} & 22.21 & \underline{84.53} & 53.46 & \underline{69.60} & \underline{55.76} \\
	& FTransUNet~\cite{ma2024ftransUNet} & TGRS’24 & 68.22 & 56.28 & 46.76 & 62.35 & 83.46 & 35.35 & 17.23 & 83.50 & 50.94 & 66.61 & 52.81 \\
	& ASANet~\cite{zhang2024asanet} & ISPRS’24 & 69.71 & 58.09 & 49.05 & 63.52 & 84.05 & 41.50 & 20.57 & 84.29 & 52.83 & 69.04 & 55.21 \\
	& MRFS~\cite{zhang2024mrfs} & CVPR’24 & 69.59 & 58.04 & \underline{49.15} & 63.78 & \textbf{84.15} & 41.26 & \underline{22.88} & 84.24 & \textbf{54.14} & 69.48 & 55.55 \\
	& {ASMFNet~\cite{ma2024asmfnet}} & {JSTARS’24} &
	{67.31} & {55.73} & {46.56} &
	{59.51} & {83.03} & {32.63} &
	{14.60} & {82.91} & {49.51} &
	{65.09} & {51.34} \\
	& {MIEFNet~\cite{fan2024miefnet}} & {GRSL’24} &
	{68.72} & {57.16} & {48.04} &
	{63.23} & {83.64} & {39.35} &
	{18.41} & {83.87} & {50.58} &
	{67.89} & {54.08} \\
	& {MFNet~\cite{ma2025mfnet}} & {TGRS’25} &
	{68.97} & {\underline{58.39}} &
	{47.78} & {63.38} & {83.88} &
	{39.51} & {20.33} & {84.02} &
	{51.86} & {68.47} & {54.61} \\
	& \cellcolor{gray!20}PAD (Ours) & \cellcolor{gray!20} & \cellcolor{gray!20}\underline{69.85} & \cellcolor{gray!20}58.10 & \cellcolor{gray!20}\textbf{49.70} & \cellcolor{gray!20}\textbf{65.39} & \cellcolor{gray!20}84.05 & \cellcolor{gray!20}\textbf{43.24} & \cellcolor{gray!20}\textbf{23.47} & \cellcolor{gray!20}\textbf{84.56} & \cellcolor{gray!20}\underline{53.93} & \cellcolor{gray!20}\textbf{70.14} & \cellcolor{gray!20}\textbf{56.26} \\
	\bottomrule[1.1pt]
	\end{tabular}
\end{table*}

\subsubsection{Evaluations on \textsc{WHU-OPT-SAR}}

As shown in \cref{table2}, PAD achieves new state-of-the-art results: 56.26\% mIoU, 84.56\% OA, 53.93\% mKappa, and 70.14\% mF1. This surpasses the best unimodal method (DANet-RGB~\cite{fu2019danet}: 53.25\% mIoU, 83.19\% OA) and the multi-modal baseline CMX~\cite{zhang2023cmx} (55.76\% mIoU, 84.53\% OA). For fine-grained classes, PAD attains 43.24\% IoU on roads and 49.70\% on villages, improving over MRFS by 1.48\% and 0.66\%, respectively. 

\begin{figure*}[!t]
	\centering
	\includegraphics[width = 0.95\linewidth]{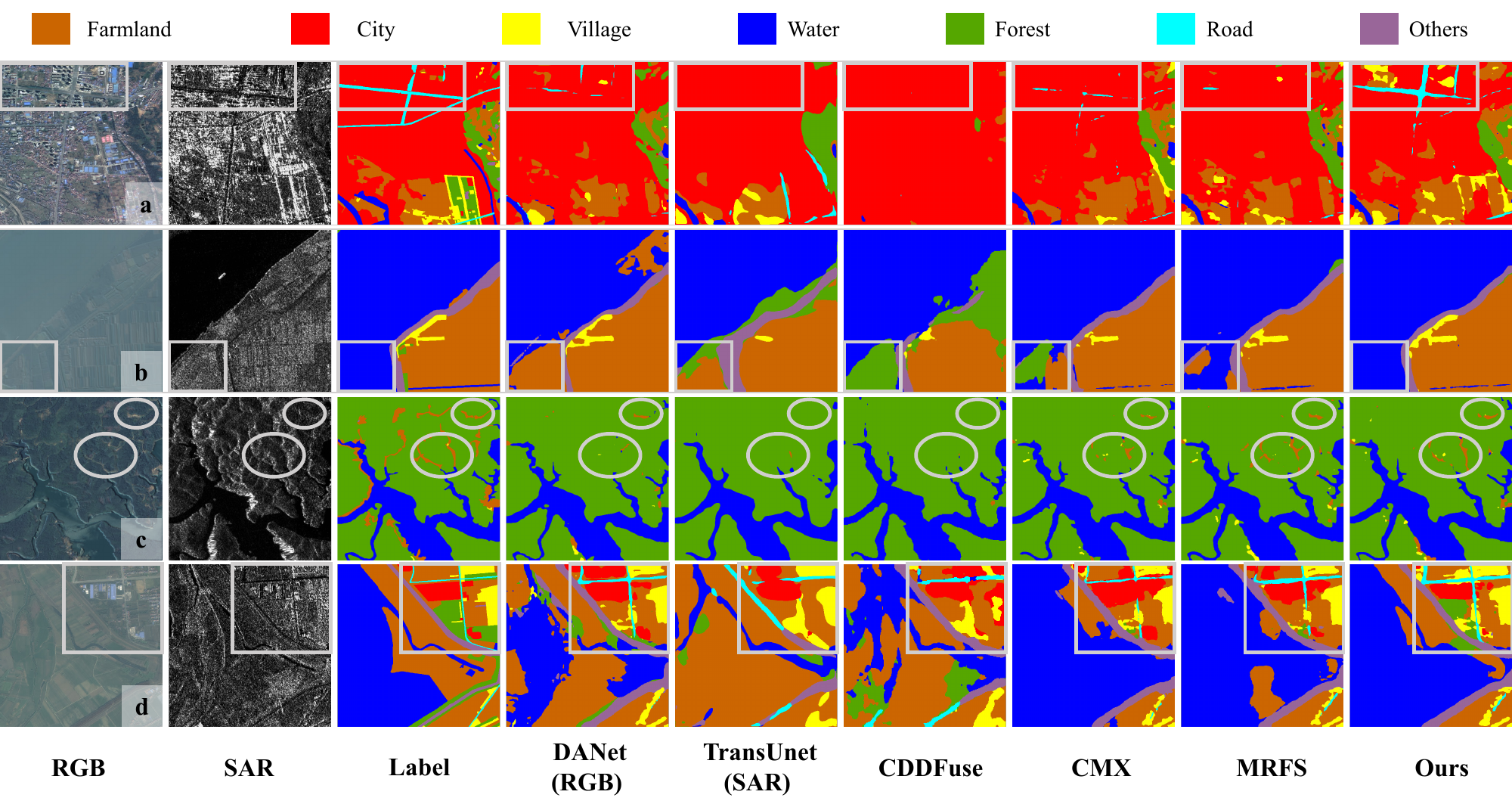}
	\caption{\textbf{Visualization Results on \textsc{WHU-OPT-SAR}.} Four representative scenarios: 
	(a) building-dense area; 
	(b) specular water bodies; 
	(c) vegetated region; and 
	(d) urban-rural transition. 
	For each scenario, images are shown (left to right): RGB, SAR, pseudo-color ground truth, and predictions from six competing methods. We compare PAD against unimodal baselines (DANet-RGB and TransUNet-SAR) and multi-modal approaches (CDDFuse, CMX, and MRFS). Regions of interest are highlighted with gray boxes (buildings/water) and circles (vegetation); see \S\ref{sec:iv_c} for details.}
	\label{fig6}
\end{figure*}

Qualitative results in \cref{fig6} further highlight PAD’s benefits. In \cref{fig6}(a), PAD recovers occluded road segments (cyan) in building-dense areas via SAR phase cues, whereas RGB-only models miss them. In \cref{fig6}(b), PAD distinguishes specular reflections from true water boundaries under cloud interference. In \cref{fig6}(c)-(d), PAD resolves vegetation spectral ambiguities and maintains consistency across urban-rural transitions.

\begin{table*}[!t]
	\centering
	\setlength{\tabcolsep}{8pt}
	\caption{\textbf{Quantitative Results (\%) on \textsc{DDHR-SK}.} Per-class IoU is reported for five land-cover categories (Buildings, Roads, Greenery, Water, and Farmland), along with OA, mKappa, mF1, and mIoU.}
	\label{table3}
	\begin{tabular}{c|ll|ccccc|cccc}
	\toprule[1.1pt]
	& \multirow{2}[2]{*}{Method} & \multirow{2}[2]{*}{Venue} & \multicolumn{5}{c|}{Per-class IoU} & \multirow{2}[2]{*}{OA} & \multirow{2}[2]{*}{mKappa} & \multirow{2}[2]{*}{mF1} & \multirow{2}[2]{*}{mIoU} \\
	\cmidrule(lr){4-8}
	& & & Buildings & Roads & Greenery & Water & Farmland & & & & \\
	\midrule
	\multirow{8}{*}{\rotatebox{90}{RGB}}
	& DANet~\cite{fu2019danet} & CVPR’19 & 94.46 & 75.11 & 91.10 & 99.03 & 90.59 & 96.23 & 73.35 & 94.57 & 90.06 \\
	& CGNet~\cite{wu2020cgnet} & TIP’20 & 80.19 & 27.57 & 72.77 & 90.47 & 69.97 & 86.16 & 33.90 & 78.76 & 68.19 \\
	& OCRNet~\cite{yuan2020ocrnet} & ECCV’20 & 90.18 & 60.49 & 84.42 & 97.87 & 82.84 & 93.17 & 58.96 & 90.26 & 83.16 \\
	& SETR~\cite{zheng2021SETR} & CVPR’21 & 94.17 & 74.10 & 91.43 & 99.18 & 90.88 & 96.22 & 73.79 & 94.49 & 89.95 \\
	& SegFormer~\cite{xie2021segformer} & NeurIPS’21 & 92.44 & 67.11 & 88.76 & 98.82 & 88.18 & 95.00 & 66.73 & 92.71 & 87.06 \\
	& Mask2Former~\cite{cheng2022masked} & CVPR’22 & 94.16 & 80.46 & 91.08 & 98.68 & 86.88 & 96.07 & 74.05 & 94.76 & 90.25 \\
	& PIDNet~\cite{xu2023pidnet} & CVPR’23 & 87.77 & 59.01 & 80.57 & 96.89 & 77.64 & 91.44 & 53.43 & 88.56 & 80.38 \\
	& TransUNet~\cite{chen2021transunet} & MIA’24 & 90.67 & 62.68 & 85.64 & 98.41 & 83.71 & 93.64 & 61.82 & 90.95 & 84.22 \\
	\midrule
	\multirow{8}{*}{\rotatebox{90}{SAR}}
	& DANet~\cite{fu2019danet} & CVPR’19 & 91.20 & 58.97 & 85.93 & 98.13 & 82.31 & 93.57 & 59.35 & 90.28 & 83.31 \\
	& CGNet~\cite{wu2020cgnet} & TIP’20 & 81.16 & 19.58 & 69.45 & 90.27 & 60.78 & 84.63 & 25.42 & 74.96 & 64.25 \\
	& OCRNet~\cite{yuan2020ocrnet} & ECCV’20 & 86.38 & 38.70 & 78.36 & 96.18 & 71.34 & 89.52 & 41.07 & 83.54 & 74.19 \\
	& SETR~\cite{zheng2021SETR} & CVPR’21 & 90.97 & 59.57 & 84.71 & 98.16 & 80.51 & 93.27 & 59.39 & 89.99 & 82.79 \\
	& SegFormer~\cite{xie2021segformer} & NeurIPS’21 & 88.81 & 46.92 & 81.99 & 97.57 & 76.75 & 91.51 & 49.55 & 86.73 & 78.41 \\
	& Mask2Former~\cite{cheng2022masked} & CVPR’22 & 91.34 & 58.52 & 86.40 & 93.29 & 80.08 & 93.20 & 62.69 & 89.49 & 81.92 \\
	& PIDNet~\cite{xu2023pidnet} & CVPR’23 & 84.37 & 36.77 & 73.28 & 94.91 & 66.91 & 87.55 & 38.63 & 81.49 & 71.25 \\
	& TransUNet~\cite{chen2021transunet} & MIA’24 & 86.56 & 38.45 & 77.56 & 96.53 & 71.30 & 89.49 & 42.21 & 83.44 & 74.08 \\
	\midrule
	\multirow{13}{*}{\rotatebox{90}{SAR \& RGB}}
	& MCANet~\cite{Li22mcanet} & JAG’21 & 83.90 & 38.06 & 75.40 & 96.04 & 69.99 & 88.07 & 41.04 & 82.54 & 72.68 \\
	& DDHRNet~\cite{ren2022ddhr} & JAG’22 & 82.76 & 29.45 & 73.78 & 95.20 & 67.40 & 87.21 & 35.24 & 79.81 & 69.72 \\
	& CEN~\cite{wang2020cen} & TPAMI’22 & 86.61 & 40.08 & 79.34 & 97.57 & 74.26 & 90.20 & 46.91 & 84.50 & 75.57 \\
	& TokenFusion~\cite{wang2022tokenfusion} & CVPR’22 & 92.28 & 65.60 & 88.05 & 98.60 & 85.74 & 94.63 & 64.71 & 92.10 & 86.05 \\
	& CMX~\cite{zhang2023cmx} & TITS’23 & 93.50 & 72.03 & 90.02 & 99.00 & 88.03 & 95.59 & 70.48 & 93.65 & 88.52 \\
	& CDDFuse~\cite{zhao2023cddfuse} & CVPR’23 & 82.19 & 35.87 & 69.95 & 93.94 & 54.50 & 85.06 & 66.26 & 78.55 & 67.29\\
	& FTransUNet~\cite{ma2024ftransUNet} & TGRS’24 & 86.78 & 40.51 & 78.70 & 97.24 & 73.13 & 89.91 & 46.02 & 84.35 & 75.27 \\
	& ASANet~\cite{zhang2024asanet} & ISPRS’24 & 86.67 & 43.18 & 79.02 & 97.28 & 73.99 & 90.17 & 46.40 & 85.03 & 76.03 \\
	& MRFS~\cite{zhang2024mrfs} & CVPR’24 & \underline{94.63} & \underline{76.20} & \underline{91.30} & \underline{99.11} & \underline{90.22} & \underline{96.32} & \underline{74.23} & \underline{94.72} & \underline{90.29} \\
	& {ASMFNet~\cite{ma2024asmfnet}} & {JSTARS’24} &
	{86.24} & {37.52} & {79.08} &
	{97.45} & {73.38} & {89.92} &
	{45.24} & {83.77} & {74.74} \\
	& {MIEFNet~\cite{fan2024miefnet}} & {GRSL’24} &
	{90.56} & {63.90} & {85.20} &
	{98.39} & {82.19} & {93.46} &
	{61.20} & {90.89} & {84.05} \\
	& {MFNet~\cite{ma2025mfnet}} & {TGRS’25} &
	{89.79} & {56.65} & {83.61} &
	{98.10} & {80.32} & {92.67} &
	{57.54} & {89.23} & {81.70} \\

	& \cellcolor{gray!20}PAD (Ours) & \cellcolor{gray!20} & \cellcolor{gray!20}\textbf{97.09} & \cellcolor{gray!20}\textbf{86.12} & \cellcolor{gray!20}\textbf{95.38} & \cellcolor{gray!20}\textbf{99.47} & \cellcolor{gray!20}\textbf{94.89} & \cellcolor{gray!20}\textbf{98.04} & \cellcolor{gray!20}\textbf{84.38} & \cellcolor{gray!20}\textbf{97.16} & \cellcolor{gray!20}\textbf{94.59} \\

	\bottomrule[1.1pt]
	\end{tabular}
\end{table*}

\subsubsection{Evaluations on \textsc{DDHR-SK}}

As shown in \cref{table3}, PAD exhibits superior performance under adverse (cloudy) conditions on \textsc{DDHR-SK}. It achieves 94.59\% mIoU and 98.04\% OA, outperforming leading unimodal methods (\textit{e.g.}, Mask2Former-RGB: 90.25\% mIoU, 94.76\% mF1; DANet-SAR: 83.31\% mIoU, 90.28\% mF1) and the multi-modal baseline MRFS (90.29\% mIoU, 74.23\% mKappa). PAD also attains 84.38\% mKappa and 97.16\% mF1. Boundary-sensitive categories achieve 86.12\% IoU on roads, surpassing MRFS by 9.92\%.

\begin{figure*}[!t]
	\centering
	\includegraphics[width = 0.95\linewidth]{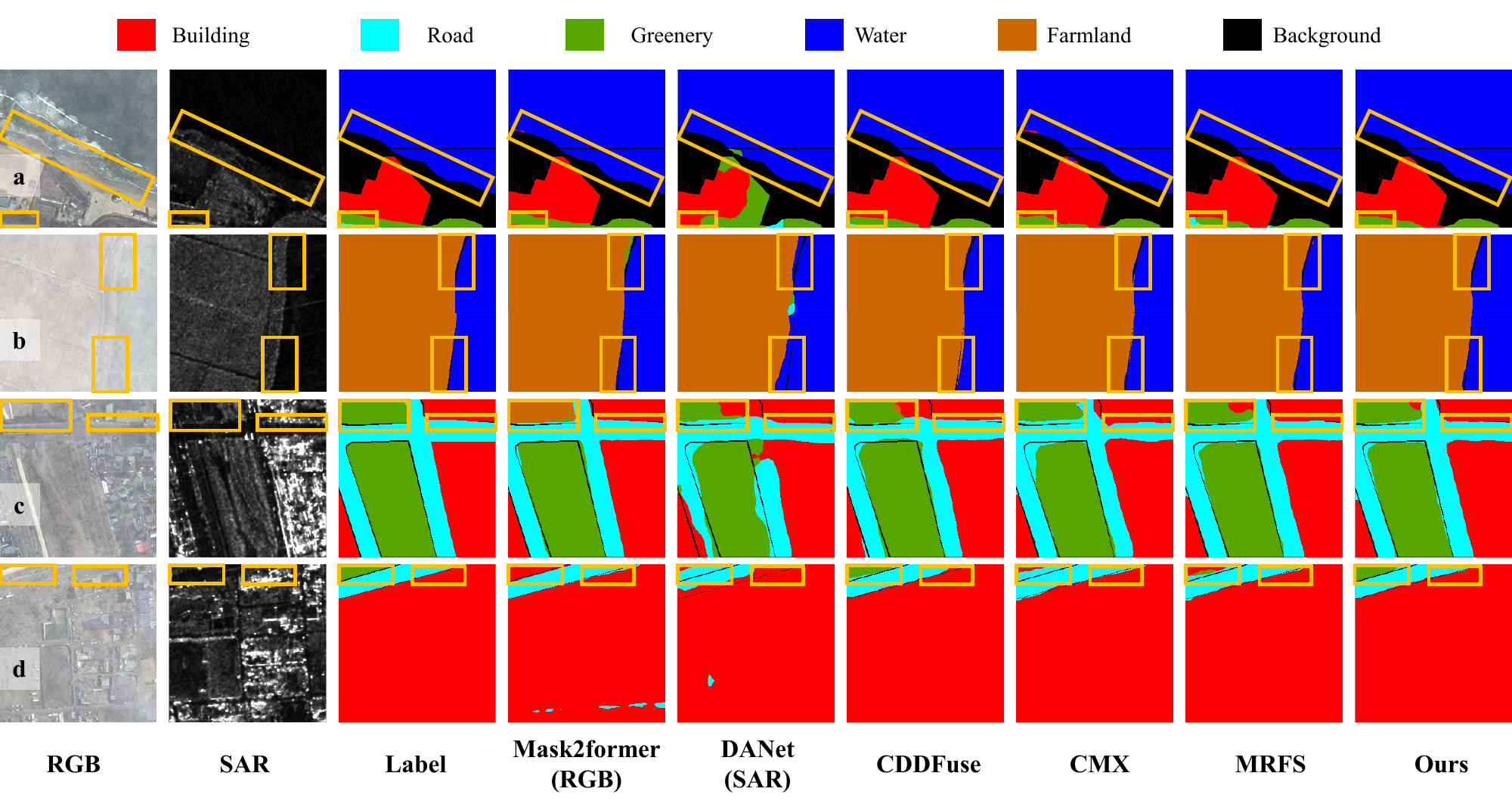}
	\caption{\textbf{Visualization Results on \textsc{DDHR-SK}.} Four boundary-focused scenarios: 
	(a) coastal demarcation; 
	(b) water-cropland interface; 
	(c) vegetation-building complex; and 
	(d) road-network topology. 
	For each scenario, panels (left to right) show the RGB input, SAR coherence map, pseudo-color ground truth, and predictions from six methods. We compare PAD against unimodal baselines (Mask2Former-RGB and DANet-SAR) and multi-modal approaches (CDDFuse, CMX, and MRFS). Critical boundary regions are highlighted with orange boxes; see \S\ref{sec:iv_c} for details.}
	\label{fig7}
\end{figure*}

Visual results demonstrate PAD’s ability for cross-modal boundary refinement. In \cref{fig7}(a), PAD accurately reconstructs periodically submerged coastlines via SAR coherence analysis, whereas RGB-only models confuse tidal sediments with terrestrial areas. In \cref{fig7}(b), PAD leverages SAR features to delineate water-infiltration boundaries in irrigated fields, overcoming the over-segmentation common in optical sensors. In \cref{fig7}(c)-(d), PAD produces continuous road networks under vegetation occlusion and precise material boundaries in vegetation-building complexes.

\subsection{Ablation and Efficiency Analysis}

\subsubsection{Ablation Study}

We conduct comprehensive ablations on \textsc{WHU-OPT-SAR} to validate the contributions of our core components: the Amplitude Spectrum Fusion (ASF) module, and the Phase Spectrum Correction (PSC) module. 

\begin{table*}[!t]
	\centering
	{
	\setlength{\tabcolsep}{9pt}
	\caption[]{\textbf{Ablation Results (\%) on Main Modules for \textsc{WHU-OPT-SAR}.} \greencheck{} and \bluecross{} indicate the inclusion or exclusion of ASF and PSC, respectively.}
	\label{table4}
	\begin{tabular}{cc|ccccccc|cccc}
	\toprule[1.1pt]
	\multicolumn{2}{c|}{Components} & \multicolumn{7}{c|}{Per-class IoU} & \multirow{2}[2]{*}{OA} & \multirow{2}[2]{*}{mKappa} & \multirow{2}[2]{*}{mF1} & \multirow{2}[2]{*}{mIoU} \\
	\cmidrule(lr){1-2} \cmidrule(lr){3-9}
	ASF & PSC & Farmland & City & Village & Water & Forest & Road & Others & & & & \\
	\midrule
	 \bluecross & \bluecross & 69.46 & 57.16 & 49.01 & 65.15 & 83.87 & 41.57 & 21.39 & 84.32 & 51.96 & 69.23 & 55.37 \\
	 \greencheck & \bluecross & 69.91 & \textbf{58.51} & \underline{49.53} & \underline{65.52} & 84.05 & 42.63 & 22.58 & 84.55 & \textbf{54.53} & 69.93 & 56.11 \\
	 \bluecross & \greencheck & \underline{69.98} & 57.32 & 49.36 & \textbf{65.62} & \textbf{84.12} & \textbf{43.42} & 23.01 & \textbf{84.57} & 53.80 & 69.98 & 56.12 \\
	\rowcolor{gray!20}
	 \greencheck & \greencheck & 69.85 & 58.09 & \textbf{49.71} & 65.39 & 84.05 & 43.24 & \textbf{23.47} & 84.56 & \underline{53.96} & \textbf{70.14} & \textbf{56.26} \\
	\bottomrule[1.1pt]
	\end{tabular}
	}
\end{table*}

\paragraph{{Main Modules}}

{Four variants (see \cref{table4}) are evaluated. The baseline, which replaces substitutes PAD (ASF+PSC) with pixel-wise addition, achieves the lowest mIoU (55.37\%). Enabling ASF alone reaches 56.11\% mIoU; PSC alone provides 56.12\% mIoU. Integrating two modules achieves the best performance (56.26\% mIoU), demonstrating effective synergy.} 

\begin{table*}[!t]
	\centering
	\setlength{\tabcolsep}{8pt}
	\caption{\textbf{Ablation on Fusion Modules (\%) for \textsc{WHU-OPT-SAR}.} Amplitude-only, phase-only, and combined (amplitude{+}phase) fusion designs are compared, each integrated into our framework.}
	\label{table5}
	\begin{tabular}{c|l|ccccccc|cccc}
	\toprule[1.1pt]
	& \multirow{2}[2]{*}{Fusion module}
	& \multicolumn{7}{c|}{Per-class IoU}
	& \multirow{2}[2]{*}{OA} & \multirow{2}[2]{*}{mKappa} & \multirow{2}[2]{*}{mF1} & \multirow{2}[2]{*}{mIoU} \\
	\cmidrule(lr){3-9}
	& & Farmland & City & Village & Water & Forest & Road & Others & & & & \\
	\midrule
	\multirow{4}{*}{\rotatebox{90}{Amplitude}}
	& FelNFN~\cite{liang2024felnfn}
	& 66.93 & 55.80 & 45.15 & 61.27 & 82.77 & 35.54 & 19.17
	& 82.95 & 47.31 & 66.46 & 52.38 \\
	& FCENet~\cite{wang2025fcenet}
	& 68.27 & 55.18 & 46.79 & 61.17 & 83.43 & 37.09 & 16.71
	& 83.43 & 49.82 & 66.52 & 52.66 \\
	& FGSwin~\cite{zhang2025fgswin}
	& \underline{69.87} & \underline{57.43} & \underline{49.03} & \underline{65.04}
	& \textbf{84.12} & \textbf{43.16} & \underline{22.26}
	& \underline{84.51} & \underline{53.53} & \underline{69.70} & \underline{55.85} \\
	& \cellcolor{gray!20}PAD (Ours)
	& \cellcolor{gray!20}\textbf{69.91} & \cellcolor{gray!20}\textbf{58.51}
	& \cellcolor{gray!20}\textbf{49.53} & \cellcolor{gray!20}\textbf{65.52}
	& \cellcolor{gray!20}\underline{84.05} & \cellcolor{gray!20}\underline{42.63}
	& \cellcolor{gray!20}\textbf{22.58}
	& \cellcolor{gray!20}\textbf{84.55}
	& \cellcolor{gray!20}\textbf{54.53} & \cellcolor{gray!20}\textbf{69.93}
	& \cellcolor{gray!20}\textbf{56.11} \\
	\midrule
	\multirow{3}{*}{\rotatebox{90}{Phase}}
	& FelNFN~\cite{liang2024felnfn}
	& 60.66 & 53.25 & 31.99 & 54.14 & 80.64 & 19.84 & 15.02
	& 79.42 & 33.32 & 58.89 & 45.08 \\
	& Channel attention
	& \underline{69.58} & \textbf{58.66} & \textbf{49.78} & \underline{64.28}
	& \underline{83.96} & \textbf{43.43} & \underline{22.65}
	& \underline{84.39} & \underline{53.54} & \underline{69.93} & \underline{56.05} \\
	& \cellcolor{gray!20}PAD (Ours)
	& \cellcolor{gray!20}\textbf{69.98} & \cellcolor{gray!20}\underline{57.32}
	& \cellcolor{gray!20}\underline{49.36} & \cellcolor{gray!20}\textbf{65.62}
	& \cellcolor{gray!20}\textbf{84.12} & \cellcolor{gray!20}\underline{43.42}
	& \cellcolor{gray!20}\textbf{23.01}
	& \cellcolor{gray!20}\textbf{84.57}
	& \cellcolor{gray!20}\textbf{53.80} & \cellcolor{gray!20}\textbf{69.98}
	& \cellcolor{gray!20}\textbf{56.12} \\
	\midrule
	\multirow{6}{*}{\rotatebox{90}{Whole}}
	& FelNFN~\cite{liang2024felnfn}
	& 66.43 & 51.97 & 44.36 & 58.48 & 82.80 & 32.33 & 12.64
	& 82.33 & 47.35 & 63.63 & 49.86 \\
	& FCENet~\cite{wang2025fcenet}
	& 66.87 & 53.01 & 46.41 & 60.30 & 82.64 & 30.21 & 14.51
	& 82.78 & 43.83 & 64.33 & 50.56 \\
	& FGSwin~\cite{zhang2025fgswin}
	& \underline{69.88} & 57.86 & \underline{49.48} & \underline{65.26}
	& \textbf{84.15} & 42.49 & 22.36
	& \textbf{84.56} & 53.41 & \underline{69.76} & \underline{55.92} \\
	& Frefusion~\cite{shi2025frefusion}
	& \textbf{69.91} & \underline{57.99} & 49.35 & 64.22 & 84.00
	& \underline{42.72} & \underline{22.46}
	& 84.41 & \textbf{53.98} & 69.69 & 55.81 \\
	& SFINet++~\cite{zhou2024SFINetplusplus}
	& 68.23 & 55.65 & 47.31 & 62.35 & 83.10 & 38.48 & 19.04
	& 83.50 & 49.56 & 67.43 & 53.45 \\
	& \cellcolor{gray!20}PAD (Ours)
	& \cellcolor{gray!20}69.85 & \cellcolor{gray!20}\textbf{58.09}
	& \cellcolor{gray!20}\textbf{49.71} & \cellcolor{gray!20}\textbf{65.39}
	& \cellcolor{gray!20}\underline{84.05} & \cellcolor{gray!20}\textbf{43.24}
	& \cellcolor{gray!20}\textbf{23.47}
	& \cellcolor{gray!20}\textbf{84.56}
	& \cellcolor{gray!20}\underline{53.96} & \cellcolor{gray!20}\textbf{70.14}
	& \cellcolor{gray!20}\textbf{56.26} \\
	\bottomrule[1.1pt]
	\end{tabular}
\end{table*}

\paragraph{Fusion Modules}

To further validate ASF and PSC, we replace each with five existing frequency-domain designs, FeINFN~\cite{liang2024felnfn}, FCENet~\cite{wang2025fcenet}, FGSwin~\cite{zhang2025fgswin}, Frefusion~\cite{shi2025frefusion}, and SFINet++~\cite{zhou2024SFINetplusplus}, while keeping the rest of our framework unchanged (see \cref{table5}). These methods represent strategies such as fixed-kernel spectral convolution (FeINFN), frequency-spatial interaction via Hadamard products/matrix multiplication (FCENet), learned spectral filtering (FGSwin), attention-based cross-domain fusion (Frefusion), and amplitude-phase parallel processing (SFINet++). Our ASF outperforms other amplitude-focused methods (56.11\% \textit{vs.} 55.85\% mIoU for FGSwin), and PSC consistently exceeds phase baselines (56.12\% \textit{vs.} 56.05\% for channel attention and 45.08\% for FeINFN). At the full-fusion level, our integrated design still leads (56.26\% mIoU), indicating that both modules are effective in isolation and synergistic when combined. These results confirm that adaptive frequency modulation (ASF) and spatially guided phase correction (PSC) offer structurally distinct and more effective alternatives to existing paradigms. 

\begin{table*}[!t]
	\centering
	\setlength{\tabcolsep}{9pt}
	\caption{\textbf{Ablation on Loss-Function Weights (\%) for \textsc{WHU-OPT-SAR}.} We vary $\lambda_1$ (with $\lambda_2$ fixed at 0.1) and $\lambda_2$ (with $\lambda_1$ fixed at 0.4).}
	\label{table6}
	\begin{tabular}{l|ccccccc|cccc}
	\toprule[1.1pt]
	\multirow{2}[2]{*}{Configuration}
	& \multicolumn{7}{c|}{Per-class IoU}
	& \multirow{2}[2]{*}{OA}
	& \multirow{2}[2]{*}{mKappa}
	& \multirow{2}[2]{*}{mF1}
	& \multirow{2}[2]{*}{mIoU} \\
	\cmidrule(lr){2-8}
	& Farmland & City & Village & Water & Forest & Road & Others & & & & \\
	\midrule
	\multicolumn{12}{c}{\textit{Fixed} $\lambda_2 = 0.1$} \\
	\midrule
	$\lambda_1 = 0.01$
	& \textbf{69.87} & 57.71 & 49.43 & 65.31 & \textbf{84.12} & \textbf{43.35} & 23.22
	& 84.54 & 53.60 & 70.02 & 56.14 \\
	\rowcolor{gray!20}
	$\lambda_1 = 0.4$
	& 69.85 & \textbf{58.09} & \textbf{49.71} & \textbf{65.39} & 84.05 & 43.24 & \textbf{23.47}
	& \textbf{84.56} & \textbf{53.96} & \textbf{70.14} & \textbf{56.26} \\
	$\lambda_1 = 1.0$
	& 69.56 & 57.30 & 49.10 & \textbf{65.39} & 83.84 & 41.59 & 20.97
	& 84.37 & 52.20 & 69.21 & 55.39 \\
	\midrule
	\multicolumn{12}{c}{\textit{Fixed} $\lambda_1 = 0.4$} \\
	\midrule
	$\lambda_2 = 0.01$
	& 69.83 & 56.80 & 48.67 & \textbf{65.63} & \textbf{84.19} & 43.17 & 22.82
	& 84.52 & 53.54 & 69.76 & 55.87 \\
	\rowcolor{gray!20}
	$\lambda_2 = 0.1$
	& \textbf{69.85} & \textbf{58.09} & \textbf{49.71} & 65.39 & 84.05 & 43.24 & \textbf{23.47}
	& \textbf{84.56} & \textbf{53.96} & \textbf{70.14} & \textbf{56.26} \\
	$\lambda_2 = 1.0$
	& 69.83 & 57.74 & 49.37 & 64.95 & 83.95 & \textbf{43.45} & 22.50
	& 84.46 & 53.87 & 69.84 & 55.97 \\
	\bottomrule[1.1pt]
	\end{tabular}
\end{table*}

\paragraph{Loss Function Weights}

We conduct weight ablations (see \cref{table6}) to assess the trade-off between the amplitude consistency loss $\mathcal{L}_{\mathrm{amp}}$ and the segmentation loss $\mathcal{L}_{\mathrm{seg}}$. The best mIoU (56.26\%) is achieved with $\lambda_1 = 0.4$ and $\lambda_2 = 0.1$. Increasing $\lambda_1$ above 0.4 degrades performance, especially on under-represented classes, whereas too small $\lambda_2$ significantly reduces accuracy, confirming that segmentation supervision must remain dominant. 

\begin{table*}[!t]
	\centering
	\setlength{\tabcolsep}{10pt}
	\caption{\textbf{Ablation on Initial Frequency Radius (\%) for \textsc{WHU-OPT-SAR}.} The trainable radius parameter in the ASF module is varied.}
	\label{table7}
	\begin{tabular}{c|ccccccc|cccc}
	\toprule[1.1pt]
	\multirow{2}[2]{*}{Radius}
	& \multicolumn{7}{c|}{Per-class IoU}
	& \multirow{2}[2]{*}{OA}
	& \multirow{2}[2]{*}{mKappa}
	& \multirow{2}[2]{*}{mF1}
	& \multirow{2}[2]{*}{mIoU} \\
	\cmidrule(lr){2-8}
	& Farmland & City & Village & Water & Forest & Road & Others & & & & \\
	\midrule
	0.01
	& \textbf{70.08} & 57.44 & 49.31 & \textbf{66.29} & \textbf{84.18} & 43.08 & 22.25
	& \textbf{84.66} & 53.92 & 69.88 & 56.09 \\
	\rowcolor{gray!20}
	0.1
	& 69.85 & \underline{58.09} & \textbf{49.71} & 65.39 & 84.05 & 43.24 & \underline{23.47}
	& 84.56 & \textbf{53.96} & \textbf{70.14} & \textbf{56.26} \\
	1
	& 69.79 & \textbf{58.19} & \textbf{49.71} & 64.81 & 84.15 & \textbf{43.44} & 23.03
	& 84.52 & 53.88 & 70.04 & 56.16 \\
	10
	& 69.71 & 57.71 & 49.69 & \underline{65.63} & \textbf{84.18} & 43.21 & \textbf{23.56}
	& \underline{84.57} & \underline{53.93} & \underline{70.12} & \underline{56.24} \\
	100
	& \underline{69.90} & 57.23 & 49.57 & 65.23 & 84.13 & \underline{43.34} & 22.94
	& 84.53 & 53.88 & 69.93 & 56.05 \\
	\bottomrule[1.1pt]
	\end{tabular}
\end{table*}

\paragraph{Initial Frequency Radius}

We vary the initial frequency radius $r$, which controls the preserved low-frequency proportion in ASF (see \cref{table7}). The optimal balance is at $r = 0.1$ (56.26\% mIoU). Values that overly suppress or over-smooth low-frequency content degrade performance, although the method remains robust for $r \in [0.1,10]$.

\begin{table}[!t]
	\centering
	\setlength{\tabcolsep}{5pt}
	\caption{\textbf{Computational Efficiency Comparison.} Model size (\#Params), per-image inference time, and throughput (FPS) for competing methods. Fewer parameters and shorter inference times, along with higher FPS, indicate better efficiency.}
	\label{table8}
	\begin{tabular}{l|ccc}
	\toprule[1.1pt]
	Method & \#Params (M) & Inference time (ms) & Throughput (FPS) \\
	\midrule
	CMX~\cite{zhang2023cmx} & 15.00 & 206.70 & 1,957.8 \\
	MRFS~\cite{zhang2024mrfs} & 10.30 & 695.85 & 1,342.2 \\
	ASANet~\cite{zhang2024asanet}& 7.82 & 72.91 & 2,245.2 \\
	\cellcolor{gray!20}{PAD (Ours)} & \cellcolor{gray!20}\textbf{3.88} & \cellcolor{gray!20}\textbf{5.16} & \cellcolor{gray!20}\textbf{3,228.7} \\
	\bottomrule[1.1pt]
	\end{tabular}
\end{table}

\subsubsection{Computational Efficiency}

We evaluate computational efficiency in terms of model size (fusion modules only), inference time per image, and frames per second (FPS), summarized in \cref{table8}. We compare against three high-mIoU methods, CMX~\cite{zhang2023cmx}, MRFS~\cite{zhang2024mrfs}, and ASANet~\cite{zhang2024asanet}. All metrics are aggregated across fusion stages with the same baseline model.

As shown in \cref{table8}, PAD uses only 3.882 M parameters, achieves the fastest inference time (5.16 ms per image), and delivers the highest FPS (3{,}228.7). These results indicate that PAD not only matches or exceeds the accuracy of existing methods but also offers real-time efficiency suitable for operational deployment.

\subsection{Further Analysis and Visualization}

\subsubsection{Visualization at the Distribution Level}

To further analyze feature distributions, we extract multi-scale features from the RGB and SAR encoders and from the fusion modules. After upsampling to the original resolution, we concatenate features channel-wise. To balance computational efficiency with discriminative power, we apply a two-stage dimensionality reduction: 
\textit{a)} random spatial sampling and 
\textit{b)} principal component analysis (PCA), retaining 50 components (over 89\% variance). 
The final visualization is generated using t-SNE~\cite{van2008tsne} (perplexity = 30, 1{,}000 iterations). 

\begin{figure*}[!t]
	\centering
	\begin{tabular}{c}
	\includegraphics[width = 0.9\linewidth]{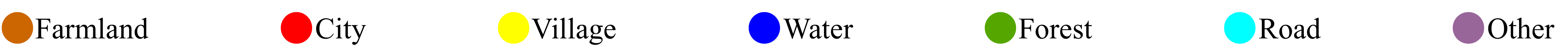}
	\vspace{-5pt} \\ 
	\end{tabular}
	\begin{tabular}{ccc}
	\includegraphics[width = 0.3\linewidth]{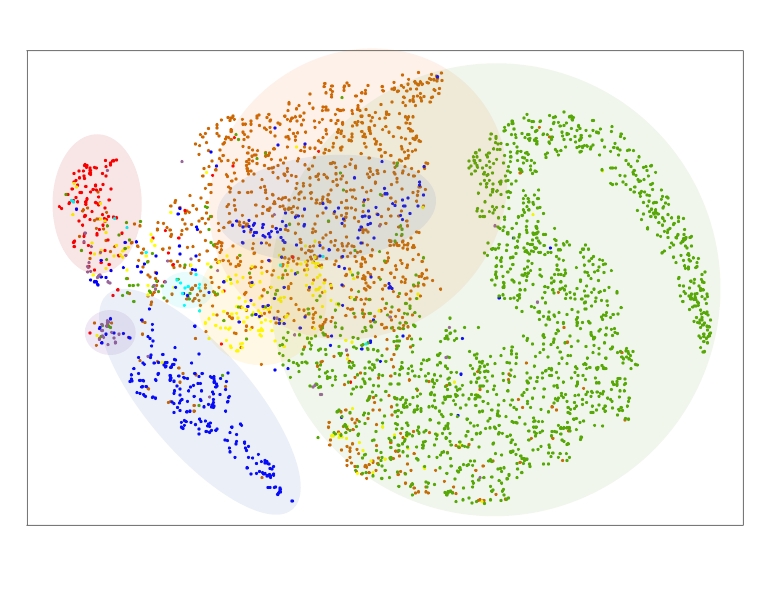} &
	\includegraphics[width = 0.3\linewidth]{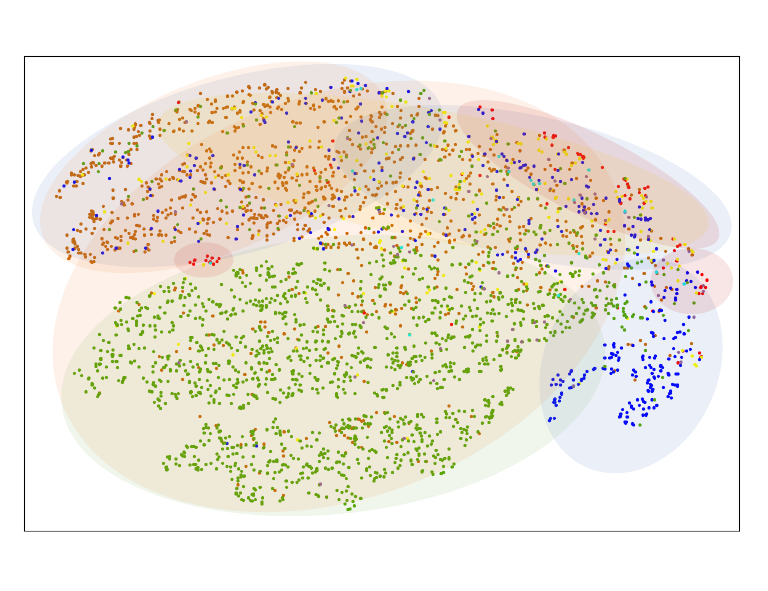} &
	\includegraphics[width = 0.3\linewidth]{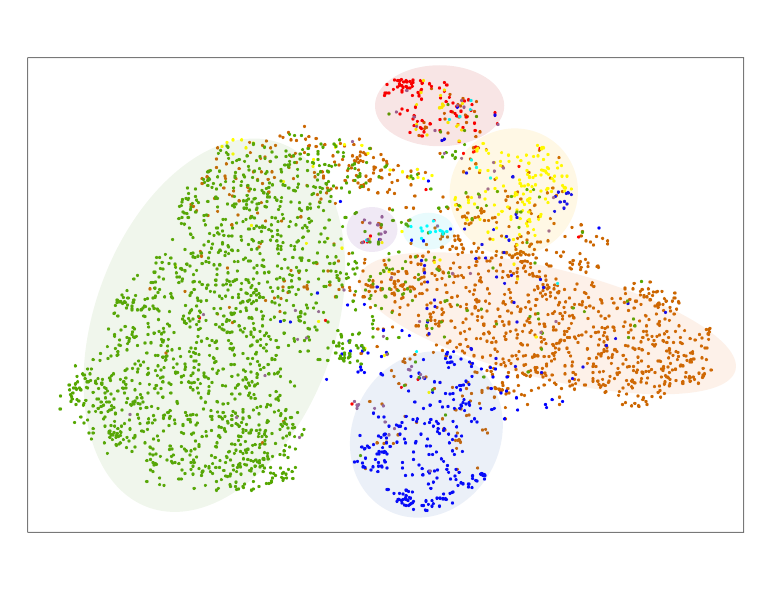}
	\vspace{-10pt} \\
	\footnotesize\textbf{(a) Before PAD fusion (RGB branch)} & \footnotesize\textbf{(b) Before PAD fusion (SAR branch)} & \footnotesize\textbf{(c) After PAD fusion}
	\end{tabular}
	\caption{\textbf{t-SNE Visualization of Cross-Modal Feature Distribution Evolution.} 
	(a) RGB-encoder features show spectral continuity but exhibit blurred class boundaries; 
	(b) SAR-encoder features capture geometric details but are affected by speckle noise; 
	(c) PAD-fused features demonstrate enhanced intra-class compactness and improved inter-class separability.}
	\label{fig8}
\end{figure*}

As shown in \cref{fig8}, PAD-fused features exhibit enhanced intra-class compactness and improved inter-class separability. Quantitatively, mean intra-class standard deviation is reduced by 20.39\% for roads, 13.16\% for water, 6.46\% for forest, and 4.43\% for farmland compared with the best unimodal baselines. Moreover, Euclidean distances between critical class pairs increase significantly: City/Village by 75.75\% ($66.56 \rightarrow 116.95$) and Farmland/Forest by 13.29\% ($107.64 \rightarrow 121.89$), indicating that PAD captures cross-modal discriminative patterns effectively.

\begin{table*}[!t]
	\centering
	{
	\setlength{\tabcolsep}{7pt}
	\caption{\textbf{Performance Comparison (\%) of Models \textit{w/} and \textit{w/o} PSC under Missing RGB Regions and Cloud Obscuration on \textsc{WHU-OPT-SAR}.}}
	\begin{tabular}{c|c|ccccccc|cccc}
	\toprule[1.1pt]
	\multirow{2}[2]{*}{Scenario} & \multirow{2}[2]{*}{Method} 
	& \multicolumn{7}{c|}{Per-class IoU} 
	& \multirow{2}[2]{*}{OA} 
	& \multirow{2}[2]{*}{Kappa} 
	& \multirow{2}[2]{*}{mF1} 
	& \multirow{2}[2]{*}{mIoU} \\ 
	\cmidrule(lr){3-9}
	& & Farmland & City & Village & Water & Forest & Road & Others 
	& & & & \\
	\midrule
	\multirow{2}{*}{Missing RGB} 
	& \textit{w/o} PSC 
	& \textbf{74.65} & 75.91 & 58.03 & 80.18 & 56.84 & 57.81 & 59.49 
	& 82.28 & 70.94 & 79.23 & 66.13 \\
	& \textit{w/} \cellcolor{gray!20}PSC 
	& \cellcolor{gray!20}\textbf{74.65} & \cellcolor{gray!20}\textbf{76.25} & \cellcolor{gray!20}\textbf{58.05} & \cellcolor{gray!20}\textbf{80.25} & \cellcolor{gray!20}\textbf{57.09} & \cellcolor{gray!20}\textbf{58.54} & \cellcolor{gray!20}\textbf{59.82} 
	& \cellcolor{gray!20}\textbf{82.35} & \cellcolor{gray!20}\textbf{71.12} & \cellcolor{gray!20}\textbf{79.42} & \cellcolor{gray!20}\textbf{66.38} \\
	\midrule
	\multirow{2}{*}{Cloud Obscuration} 
	& \textit{w/o} PSC 
	& 75.33 & 75.47 & 57.42 & 76.31 & 84.30 & 53.40 & 44.94 
	& 86.90 & 63.06 & 79.23 & 66.74 \\
	& \textit{w/} \cellcolor{gray!20}PSC 
	& \cellcolor{gray!20}\textbf{75.37} & \cellcolor{gray!20}\textbf{75.73} & \cellcolor{gray!20}\textbf{57.56} & \cellcolor{gray!20}\textbf{76.47} & \cellcolor{gray!20}\textbf{84.32} & \cellcolor{gray!20}\textbf{53.68} & \cellcolor{gray!20}\textbf{45.30} 
	& \cellcolor{gray!20}\textbf{86.95} & \cellcolor{gray!20}\textbf{63.52} & \cellcolor{gray!20}\textbf{79.37} & \cellcolor{gray!20}\textbf{66.92} \\
	\bottomrule[1.1pt]
	\end{tabular}
	}
	\label{table9}
\end{table*}

\subsubsection{{Robustness of PSC under Challenging Scenarios}}

{To further evaluate the robustness of PSC beyond normal distributions, we conducted targeted ablation experiments under extreme long-tail conditions, including missing RGB regions and cloud obscuration. Results are summarized in \cref{table9}.}

\paragraph{{Missing Regions}}

{When large portions of the RGB modality are absent, models without PSC accumulate residual phase errors, particularly in categories with fine spatial structures (\textit{e.g.}, forest, road). With PSC, alignment remains stable and yields consistent improvements across macro-level metrics (\textit{e.g.}, +0.19 mF1, +0.25 mIoU).}

\paragraph{{Cloud Obscuration}}

{Under cloud occlusion, traditional co-registration often provides weak or misleading guidance. PSC mitigates this issue by preserving phase consistency, resulting in systematic improvements in category-wise IoU (\textit{e.g.}, city, village) as well as overall agreement scores.}

{These results confirm that PSC not only aligns with physical and statistical expectations but also maintains robust performance under adverse long-tail scenarios, where purely physical models typically degrade.}

\begin{table}[!t]
	\centering
	\setlength{\tabcolsep}{12pt}
	\caption{\textbf{Statistical Comparison between Models \textit{w/} and \textit{w/o} ASF.}}
	\label{table10}
	\begin{tabular}{l|cc}
	\toprule[1.1pt]
	Metric & \cellcolor{gray!20}\textit{w/} ASF & \textit{w/o} ASF \\
	\midrule
	Mean Pearson correlation & \cellcolor{gray!20}0.0364 & 0.0369 \\
	Mean Spearman correlation & \cellcolor{gray!20}0.0482 & 0.0488 \\
	\midrule
	Paired $t$-test $p$-value & \multicolumn{2}{c}{0.0016} \\
	Wilcoxon test $p$-value & \multicolumn{2}{c}{$6.4\times10^{-7}$} \\
	Cohen’s $d$ (effect size) & \multicolumn{2}{c}{-0.038} \\
	\bottomrule[1.1pt]
	\end{tabular}
\end{table}

\subsubsection{Noise Sensitivity of ASF}

To assess whether ASF amplifies high-frequency speckle noise in SAR images, we perform a correlation analysis using the Equivalent Number of Looks (ENL), a standard measure of speckle. For each image, we compute pixel-wise ENL maps and then measure Pearson and Spearman correlations between ENL and classification accuracy. As shown in \cref{table10}, the model with ASF exhibits nearly identical correlation coefficients to the baseline; differences in mean correlation are below 0.001, and Cohen’s $d = -0.038$ indicates a negligible effect size. These results confirm that ASF does not increase noise sensitivity and remains robust in low-SNR regions.

Overall, these findings confirm that PAD significantly improves classification accuracy and prediction consistency by integrating multi-modal information, suppressing noise, and preserving discriminative spectral and geometric characteristics inherent in the data.

\section{Conclusion}

In this paper, we proposed the Phase-Amplitude Decoupling (PAD) framework for fusing SAR and RGB modalities in land-cover classification. 
Our frequency-domain analysis quantitatively demonstrates that phase components remain consistent across modalities, whereas amplitude spectra capture modality-complementary details, providing the first physically grounded prior for multi-source remote-sensing fusion. 
PAD decouples and aligns phase features via the Phase Spectrum Correction (PSC) module and adaptively fuses amplitude components using the Amplitude Spectrum Fusion (ASF) module, effectively leveraging SAR’s high-frequency details together with RGB’s low-frequency context. 
Extensive experiments on two challenging datasets show that PAD outperforms existing methods in both accuracy and robustness.

Despite its demonstrated effectiveness, PAD assumes well-aligned input modalities and relies on hard-label supervision. 
In future work, we will integrate self-supervised cross-modal registration refinement and soft-label modeling, and investigate transformer-based frequency fusion to further enhance global context awareness.

\bibliographystyle{IEEEtran}
\bibliography{PAD}

\vspace{-11pt}
\begin{IEEEbiography}[{\includegraphics[width = 1in,height = 1in,clip,keepaspectratio]{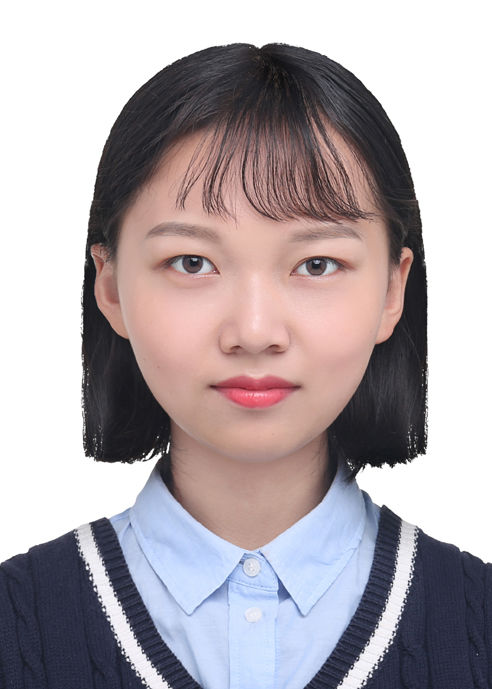}}]{Huiling Zheng}
received the B.S. degree in marine science from Shanghai Ocean University, Shanghai, China, in 2019. She is currently pursuing the M.S. degree in computer science with the School of Computer Science and Artificial Intelligence, Wuhan University of Technology, Wuhan, China. Her research interests include remote sensing image processing and AI oceanography.

\end{IEEEbiography}

\vspace{-11pt}
\begin{IEEEbiography}[{\includegraphics[width = 1in,height = 1in,clip,keepaspectratio]{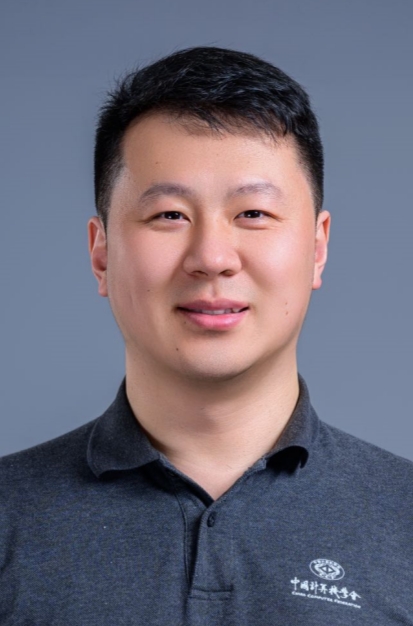}}]{Xian Zhong}
(Senior Member, IEEE) received the B.S. degree in computer science from Wuhan University, Wuhan, China, in 2007 and the Ph.D. degree in computer science from Huazhong University of Science and Technology, Wuhan, China, in 2013. He was a Visiting Scholar at Peking University, Beijing, China, 2021-2022, and at ROSE@EEE, Nanyang Technological University (NTU), Singapore, 2024. He is a Professor with the School of Computer Science and Artificial Intelligence, Wuhan University of Technology. He serves as an Associate Editor for \textit{TVC} and as an Area Chair for IJCAI, ICASSP, ICME, ICIP, and IJCNN. His research interests include multi-modal information processing, neuromorphic vision, and intelligent transportation systems.

\end{IEEEbiography}

\vspace{-11pt}
\begin{IEEEbiography}[{\includegraphics[width = 1in,height = 1in,clip,keepaspectratio]{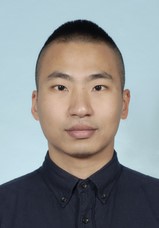}}]{Bin Liu}
(Member, IEEE) received the B.S. degree in information engineering and the M.S. and Ph.D. degrees in signal and information processing from Shanghai Jiao Tong University, Shanghai, China, in 2007, 2009, and 2015, respectively. He was a Visiting Ph.D. Student with the Dept. of Signal and Image Processing, Télécom ParisTech, Paris, France, 2012-2013. He was an Assistant Researcher with the Shanghai Key Lab of Intelligent Sensing and Recognition, Shanghai Jiao Tong University, 2015-2018, and a Postdoctoral Fellow with the College of Marine Sciences, Shanghai Ocean University, 2018-2020, where he is currently an Associate Professor and serves as the Deputy Director of the Department of Marine Technology. His research interests include AI oceanography, SAR and PolSAR image processing and understanding, image segmentation/classification, and AI-empowered fishery forecasting.

\end{IEEEbiography}

\vspace{-11pt}
\begin{IEEEbiography}[{\includegraphics[width = 1in,height = 1in,clip,keepaspectratio]{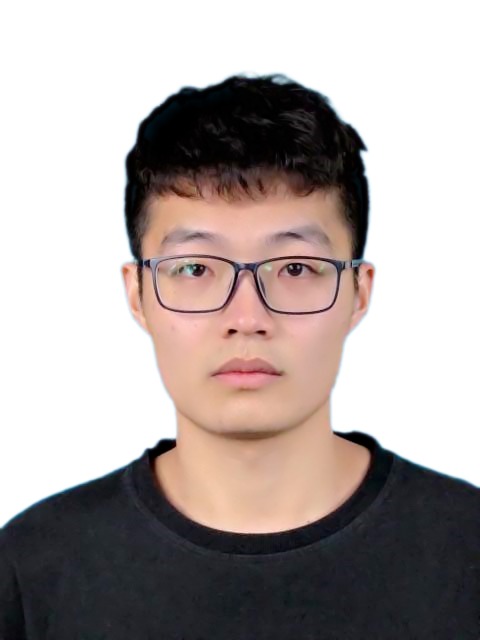}}]{Yi Xiao}
(Member, IEEE) received the B.S. degree from the School of Mathematics and Physics, China University of Geosciences, Wuhan, China, in 2020 and the M.S. and Ph.D. degrees from Wuhan University, Wuhan, China, in 2022 and 2025, respectively. He is currently with the School of Computer and Artificial Intelligence, Zhengzhou University, Zhengzhou, China. His research interests include remote sensing image/video processing and computer vision. More details are available at \url{https://xy-boy.github.io}.

\end{IEEEbiography}

\vspace{-11pt}
\begin{IEEEbiography}[{\includegraphics[width = 1in,height = 1in,clip,keepaspectratio]{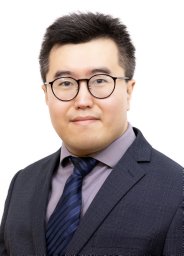}}]{Bihan Wen}
(Senior Member, IEEE) received the B.Eng. degree from Nanyang Technological University (NTU), Singapore, in 2012 and the M.S. and Ph.D. degrees from the University of Illinois at Urbana-Champaign (UIUC), Urbana-Champaign, IL, USA, in 2015 and 2018, respectively. He is a Nanyang Assistant Professor with the School of Electrical and Electronic Engineering, NTU. His honors include the NTU Early Career Teaching Excellence Award 2022, NTU Inspirational Mentor for Koh Boon Hwee Award 2021, PEB Gold Medal 2012, Best Paper Runner-Up at IEEE ICME 2020, Best Paper at IEEE ICIEA 2023, and Best Paper at IEEE MIPR 2023. He has been listed among Stanford University’s World’s Top 2\% Scientists (AI) in 2021 and 2023 and was a 2023 CASS VSPC Rising Star Runner-Up. He is an Associate Editor for \textit{TCSVT} and a Guest Editor for \textit{SPM} and \textit{JSTSP}. His interests include machine learning, computational imaging, computer vision, image processing, and AI security.

\end{IEEEbiography}

\vspace{-11pt}
\begin{IEEEbiography}[{\includegraphics[width = 1in,height = 1in,clip,keepaspectratio]{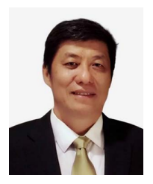}}]{Xiaofeng Li}
(Fellow, IEEE) received the B.S. degree in optical engineering from Zhejiang University, Hangzhou, China, in 1985, the M.S. degree in physical oceanography from the First Institute of Oceanography, Qingdao, China, in 1992, and the Ph.D. degree in physical oceanography from North Carolina State University, Raleigh, NC, USA, in 1997. From 1997 to 2019, he was with the National Oceanic and Atmospheric Administration (NOAA), Washington, DC, USA, working on operational satellite ocean remote sensing products. He is currently with the Institute of Oceanology, Chinese Academy of Sciences, Qingdao, China. His interests include satellite/AI oceanography, big data, and image processing. He is on the editorial boards of \textit{IJDE}, \textit{Big Earth Data}, and \textit{JOL}, and serves as an Associate Editor for \textit{TGRS} and \textit{IJRS}, as well as the Executive Editor-in-Chief for the \textit{J. Remote Sens.}.

\end{IEEEbiography}

\end{document}